\documentclass{article}
\def\arxivversion{1}
\ifdefined\arxivversion\else
    \documentclass{article}
\fi

\usepackage{iclr2027_conference,times}

\usepackage{amsmath,amsfonts,bm}

\def\eqref#1{equation~\ref{#1}}

\def\1{\bm{1}}

\DeclareMathAlphabet{\mathsfit}{\encodingdefault}{\sfdefault}{m}{sl}
\SetMathAlphabet{\mathsfit}{bold}{\encodingdefault}{\sfdefault}{bx}{n}

\usepackage[utf8]{inputenc}
\usepackage[T1]{fontenc}
\usepackage{hyperref}
\usepackage{url}
\usepackage{booktabs}
\usepackage{amsfonts}
\usepackage{amssymb}
\usepackage{amsmath}
\usepackage{graphicx}
\usepackage{makecell}
\usepackage{microtype}
\usepackage{wrapfig}
\usepackage[section]{placeins}
\usepackage{xcolor}
\usepackage{colortbl}
\usepackage{soul}
\usepackage{enumitem}

\newcommand{\eg}{\textit{e.g.}}

\newcommand{\resti}{\textsc{ReSTI}}
\newcommand{\revsitiny}{\textsc{ReVSI-Tiny}}
\newcommand{\restitiny}{\textsc{ReSTI-Tiny}}

\newcommand{\methodname}{\textsc{CROSS}}
\definecolor{benchmarkgain}{RGB}{34,110,65}
\newcommand{\benchmarkgain}[2]{#1\,{\tiny\textcolor{benchmarkgain}{(+#2)}}}
\providecommand{\pz}[1]{}
\newcommand{\finishwrap}{%
    \par
    \ifnum\value{WF@wrappedlines}>1
        \vspace{\dimexpr\value{WF@wrappedlines}\baselineskip-\baselineskip\relax}%
    \fi
    \WFclear
}

\title{From Reasoning Failures to Composable Video Spatial Intelligence}

\ifdefined\arxivversion
    \author{%
        \begin{minipage}[t]{0.96\textwidth}
            \centering
            \textbf{Pengzhan Sun$^{1}$, Junbin Xiao$^{2,\ast}$, Ramanathan Rajaraman$^{1}$, Shiu-hong Kao$^{1}$, Angela Yao$^{1}$}\\[0.5em]
            \normalfont\small
            $^{1}$National University of Singapore \quad $^{2}$University of Science and Technology of China\\[0.4em]
            {\scriptsize\ttfamily
                \{pengzhan,ayao\}@comp.nus.edu.sg \quad
                junbinxiao@ustc.edu.cn
            }
        \end{minipage}%
    }
    \iclrfinalcopy
    \hypersetup{%
        pdfauthor={Pengzhan Sun, Junbin Xiao, Ramanathan Rajaraman, Shiu-hong Kao, Angela Yao},
        pdftitle={From Reasoning Failures to Composable Video Spatial Intelligence}%
    }
\else
    \author{Anonymous Authors}
\fi

\begin{document}

\maketitle
\ifdefined\arxivversion
    \lhead{Preprint}
    \begingroup
        \renewcommand{\thefootnote}{\fnsymbol{footnote}}
        \footnotetext[1]{Corresponding author}
    \endgroup
\fi

\begin{abstract}

Spatial reasoning benchmarks evaluate vision-language models across diverse tasks, but task-level scores do not reveal which underlying capabilities account for success or failure.  Each task requires recovering spatial evidence, representing geometry, and reasoning over it.
We disentangle these capabilities by comparing predicted and ground-truth spatial context under a shared schema and coordinate contract. This comparison reveals four recurring sources of error: inaccurate perception, missing information in the spatial context, selection of the wrong measurement, and errors in reference frames or in tracking position and orientation.
Guided by this diagnosis, we develop CROSS, a training-free library of typed geometric operators and spatial skills that function over available evidence to support reliable video spatial reasoning. The resulting library supplies verified context to non-coding VLMs or callable skills to a SpatialClaw agent. We evaluate \methodname{} on five benchmarks. \methodname{} raises the average score from 55.9\% to 60.2\% on ReVSI and improves the SpatialClaw result from 62.8\% to 66.3\% on DSI-Bench.  These gains demonstrate that explicit handling of spatial conventions can repair systematic reasoning failures without additional training.

\end{abstract}

\section{Introduction}
\label{sec:intro}

Video and multi-view spatial reasoning benchmarks \citep{yang2025thinking,li2025stibench,lin2025mmsi,zhang2025dsibench,wang2026ucsbench,zhang2026revsi,wang2025mindcube,zhou2025reason4d,lin2025ostbench,wei2026streaming} evaluate models on tasks such as object geometry, distance comparison, egocentric direction, and route planning.
Each task requires models to recover spatial evidence and use appropriate quantities, reference frames, and temporal states.
Task-level scores, however, do not reveal which capabilities account for success or failure.
A wrong answer may result from missing or noisy perception, a representation that does not support the requested quantity, or misuse of otherwise sufficient evidence.
Treating these causes as a single ``spatial reasoning error'' obscures both where models struggle and which interventions could help.

Existing approaches support video spatial reasoning through large-scale training with tailored data \citep{chen2024spatialvlm, yang2026cambrian,cai2025scaling, yang2025vst,liao2025r1zero,huang2025threedr1,ouyang2025spacer,wang2025svqar1,li2025videochatr1,zhao2025embodiedr,zhou2025roborefer,wang2025vagen,lu2026vista,chen2025spacetools}, 
structured memory \citep{wu2024mindseye,yang2025thinking, chen2026egomind,ma2026blueprints}, 
scene traces \citep{wang2026ucsbench,he2026farm,linghu2026scenecot}, 
and executable code \citep{cho2026spatialclaw},
though improved task performance does not establish which underlying capabilities these approaches address.
More recent studies \citep{ropero2026riemind,yeung2026split,li2026crisp} explicitly decouple visual perception from reasoning, and demonstrate that substantial gains can be achieved by supplying perfect perceptual results.
However, perception and reasoning each encompass several distinct capabilities, and attributing failure to either alone offers limited guidance on which evidence or computation needs to improve. 

We therefore give a closer examination. We investigate the failure cases along different reasoning dimensions across both static and dynamic video spatial reasoning tasks, linking errors to specific evidence requirements and computations, and explaining why better perception resolves some tasks but leaves others unsolvable.
Specifically, by comparing video-only inference with inference augmented by spatial context, we reveal four recurring sources of error:
First, inaccurate perception limits the recovery of relevant scene information.
Second, even accurate spatial context can omit necessary information, such as room footprints or motion between stored trajectory poses.
Third, models select the wrong measurement, such as computing center-to-center instead of the requested closest-surface gap.
Fourth, models use the wrong reference frame or state, leading to left--right reversals, incorrect coordinate transformations, or reuse of the outdated pose.

Guided by these diagnoses, we develop \methodname{}, a library of \textbf{C}omposable
\textbf{R}easoning \textbf{O}perators and \textbf{S}patial
\textbf{S}kills for video spatial reasoning. Its operators perform individual computations, such as measuring a surface distance, transforming coordinates, or updating heading. We further compose these operators into task-specific skills. For example, a relative-direction skill constructs the observer's reference frame, projects the target into that frame, and determines its direction. A route planning skill also uses these computations, while updating position and heading after each step. Explicit input requirements and validity checks help detect when the available evidence cannot support a reliable result.
Explicit geometric operations already support spatial reasoning in Map2Thought \citep{gao2026map2thought}.
Yet tool access and code execution do not eliminate geometric errors: SpatialClaw reports failures in coordinate, distance, and angle computations even with sound perceptual evidence \citep{cho2026spatialclaw}.
\methodname{} targets these recurring computations, guided by the wrong-measurement and frame/state errors in our diagnosis (Section~\ref{sec:failure_taxonomy}).

We evaluate \methodname{} through two interfaces: derived spatial contexts for non-coding VLMs (\eg, Qwen3-VL \citep{bai2025qwen3} and Gemma-4 \citep{team2026gemma} ), and callable spatial skills for code agent SpatialClaw \citep{cho2026spatialclaw}.
Our experiments span five benchmarks, with controlled diagnosis covering 13 benchmark-defined subtasks in \revsitiny{} and \restitiny{}.
Beyond these diagnostic subsets, \methodname{} raises the ReVSI average from 55.9\% to 60.2\% over direct video inference \citep{zhang2026revsi}.
Crucially, we also test cross-benchmark transfer: with the operator library frozen, adding \methodname{} raises matched SpatialClaw accuracy from 62.8\% to 66.3\% on DSI-Bench \citep{zhang2025dsibench}.
This supports reuse of the operator basis through task-specific compositions, even in an agent already equipped with perception tools and code execution.

Our primary contributions are summarized below:
\begin{itemize}[leftmargin=*,topsep=2pt,itemsep=1pt]
    \item {A controlled diagnosis of video spatial reasoning across 13 dimensions that identifies four recurring sources of error: inaccurate perception, missing information in spatial context, selection of the wrong measurement, and errors in reference frames or spatial state.}
    \item {CROSS, a training-free library that turns these diagnoses into reusable geometric operators and task-specific skills.}
    \item {
    \methodname{} raises the ReVSI average from 55.9\% to 60.2\%, and the same operators raises SpatialClaw accuracy from 62.8\% to 66.3\% on DSI-Bench.}
\end{itemize}


\section{Related Work}
\label{sec:related}

\textbf{Spatial Benchmarks and Diagnostic Studies}
Spatial benchmarks span image perception in BLINK \citep{fu2024blink}, viewpoint-dependent localization in ViewSpatial-Bench \citep{li2025viewspatial}, and video-based scene reasoning in VSI-Bench \citep{yang2025thinking}.
SIBench organizes a broader evaluation around spatial perception, understanding, and planning \citep{yu2025sibench}.
ReVSI \citep{zhang2026revsi} further examines annotation validity and frame-dependent answerability.
STI-Bench \citep{li2025stibench}, DSI-Bench \citep{zhang2025dsibench}, and UCS-Bench \citep{wang2026ucsbench} extend evaluation to motion and persistent spatial memory.
End-to-end results alone do not locate failures, motivating interventions that separate perception from reasoning.
\resti{} \citep{sun2026restisourcegroundedauditrepair} audits STI-Bench against source annotations and repairs coordinate, timestamp, and target-definition errors.
RieMind \citep{ropero2026riemind} measures reasoning with a ground-truth (GT) 3D scene graph and geometric tools on static VSI-Bench; CRISP \citep{li2026crisp} contrasts predicted and GT scene-graph inputs to diagnose grounding and reasoning in static images.
SPLIT \citep{yeung2026split} replaces estimated perception-tool outputs with GT measurements while retaining a planner that composes them, revealing substantial perceptual headroom.
These studies already combine diagnostic interventions with reasoning mechanisms.
Our emphasis is on identifying recurring wrong-measurement, frame/state, and missing-information failures under a shared video-context schema across static and dynamic tasks, then testing reusable repairs through both context augmentation and code-writing agents.

\begin{figure}[t]
    \centering
    \includegraphics[width=\linewidth,trim=15bp 28bp 18bp 10bp,clip]{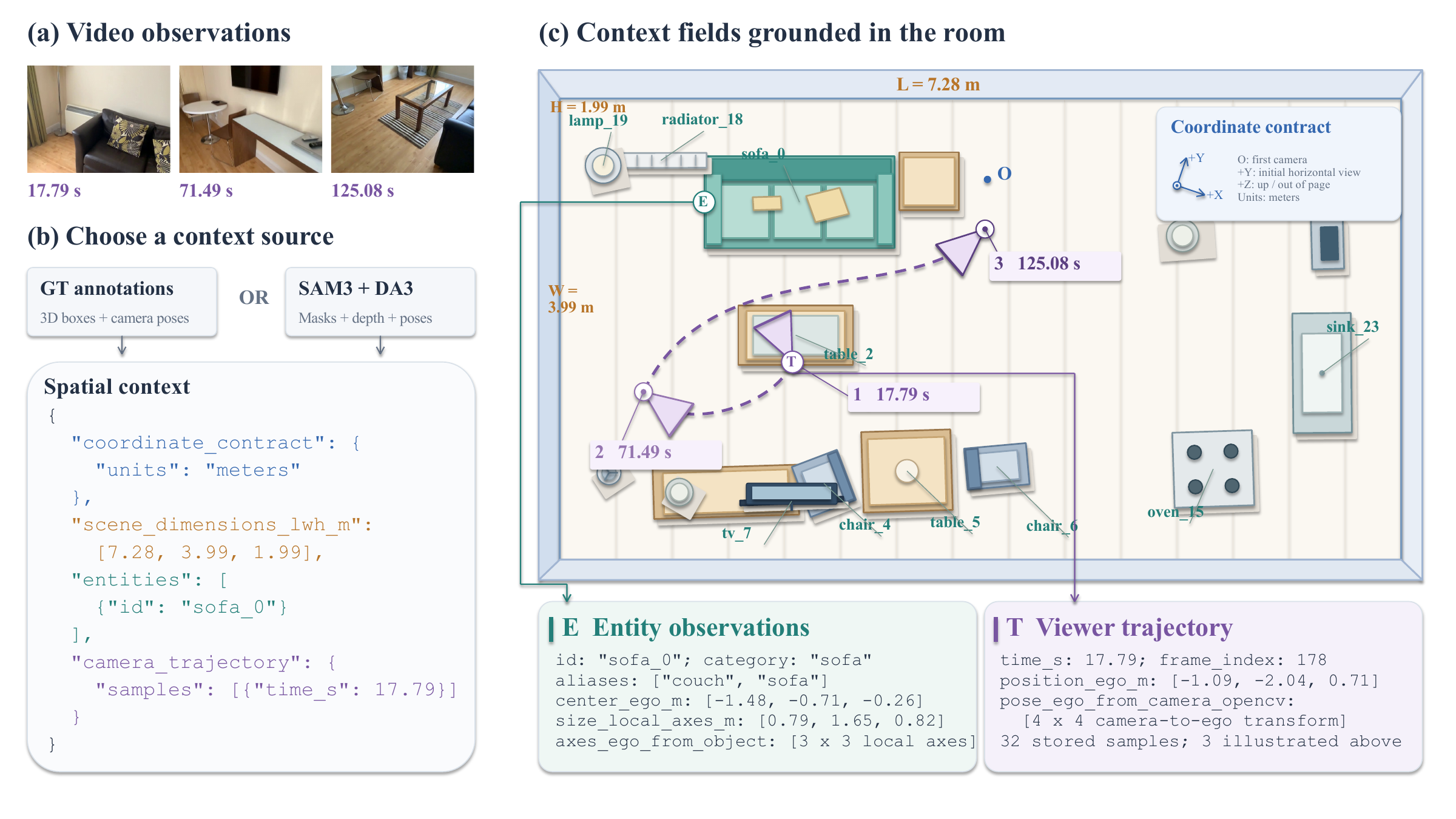}
    \caption{Illustration of spatial context.
    For each video, either ground-truth or tool-derived spatial clues
    populate the coordinate contract, scene geometry, entity
    observations, and viewer trajectory.}
    \label{fig:structured_spatial_context}
    \vspace{-3mm}
\end{figure}

\textbf{Methods for Video Spatial Reasoning}
Beyond the training approaches discussed in Section~\ref{sec:intro}, explicit representations expose spatial evidence to the reasoner.
SpatialMind \citep{zhang2025spatialmind}, See\&Trek \citep{li2025seetrek}, and TRACE \citep{hua2026trace} use structured prompts to organize scene or trajectory information.
EgoMind builds linguistic scene graphs across frames \citep{chen2026egomind}, while GR3D connects image-marked object identities to textual 3D geometry \citep{yuan2026gr3d}.
Cog3DMap maintains geometrically grounded memory tokens \citep{gwak2026cog3dmap}, whereas CoCoSI uses collaborating agents to construct cognitive maps \citep{zhang2026cocosi}.
Map2Thought \citep{gao2026map2thought} combines metric maps with deterministic geometric operations.
Agentic methods integrate specialist tools: ViSRA \citep{mou2026visra} and S-Agent \citep{dai2026sagent} organize spatial tools, VADAR synthesizes a dynamic Python API \citep{marsili2025vadar}, and SpatialClaw supplies a stateful Python execution interface \citep{cho2026spatialclaw}.
In contrast, \methodname{} derives its operators from the failures diagnosed in Section~\ref{sec:failure_taxonomy}: each targets a recurring wrong-measurement or frame/state error and makes the corresponding computation explicit, reusable, and verifiable.

\section{Diagnosing Reasoning Failures}
\label{sec:diagnostic_main}


\subsection{Problem formulation}

Let $x$ denote a sampled video, $q$ a spatial question, $c$ a structured
spatial context, and $y$ the benchmark answer.
We refer to supplying $c$ alongside the video and question as \emph{spatial
context augmentation}.
Ordinary end-to-end evaluation observes only $f(x,q)$, so an incorrect answer
does not identify whether the relevant evidence was never recovered or was
used incorrectly.
We intervene on the available evidence by comparing video-only inference
$f(x,q)$ with $f(x,q,c_{\mathrm{pred}})$ and $f(x,q,c_{\mathrm{gt}})$,
where $c_{\mathrm{pred}}$ and $c_{\mathrm{gt}}$ are predicted and ground-truth
instances of the same structured spatial context.
They share the same schema, semantic field definitions, coordinate contract,
and serialization; only the source and accuracy of their contents differ.
Comparing performance under $c_{\mathrm{gt}}$ and $c_{\mathrm{pred}}$
measures the penalty associated with estimating the same context from video.
We call this difference the \emph{perception penalty}.
On examples that pass the sufficiency and validity checks in
Appendix~\ref{sec:diagnostic_validity}, failures under
$c_{\mathrm{gt}}$ define the \emph{reasoning residual}.

\subsection{Structured spatial context}
\label{sec:spatial_context}

We call the injected artifact \emph{structured spatial context} because it
contains externally constructed scene evidence rather than model-generated
reasoning steps.
For a video-question pair, we represent it as
$c=\{\kappa,\,g,\,p_{1:T},\,\mathcal{E}\}$.
The \textbf{coordinate contract} $\kappa$ declares the world and camera
frames, axis directions, handedness, units, and pose-transform direction.
The \textbf{global geometry} $g$ summarizes the metric extent 
of the observed scene or queried region.
The \textbf{viewer trajectory} $p_{1:T}$ records timestamped camera positions
and poses, including the endpoints of the queried interval.
The \textbf{entity observations} $\mathcal{E}=\{e_i\}$ assign each persistent
entity a category, 3D position, and physical extent.
All geometric fields use a common ego-anchored, gravity-aligned coordinate
frame; its axis and transformation conventions are detailed in the
supplementary material.
Figure~\ref{fig:structured_spatial_context} illustrates how the two evidence
sources populate this shared representation.

GT context populates this schema from source annotations, whereas
predicted context uses SAM3 entity masks and tracks together with DA3 metric
geometry and camera poses.
We retain only information types available in both branches and apply the same
coordinate normalization and timestamp schedule.
At inference, $c$ is serialized as a separate context block alongside the
video and question; it contains observations and their conventions, but no
model-generated chain of thought, candidate-dependent calculation, or final
answer.
All arms share the question, answer parser, and frame budget; the augmented arms share the context renderer, and provenance fields are stripped.

\subsection{Failure taxonomy}
\label{sec:failure_taxonomy}

\begin{wrapfigure}{R}{0.5\textwidth}
  \vspace{-\intextsep}
  \centering
  \includegraphics[width=\linewidth]{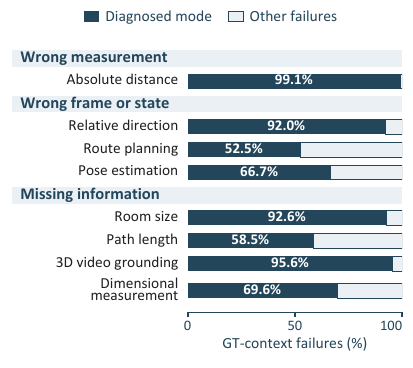}
  \small
  \caption{Share of GT-context failures matching the diagnosed mode (dark);
  pale segments show other failures.
  Rules and counts: Appendix~\ref{sec:gt_context_audit}.}
  \label{fig:gt_residual_breakdown}
\end{wrapfigure}
Our controlled comparison identifies four recurring failure types.
First, comparing predicted and GT context in Table~\ref{tab:diagnostic}
reveals a \textbf{perception gap}.
Beyond perception, almost all absolute-distance failures under GT context
match center-to-center distance within 5\%, indicating use of the
\textbf{wrong measurement} instead of the requested closest-surface distance.
A third failure type is using the \textbf{wrong frame or state} when
interpreting spatial relations or motion.
This manifests as observer-frame errors in relative direction, left--right
turn reversals in route planning, and incorrect frame composition or reuse of
the initial pose in pose estimation.
Finally, \textbf{missing information} limits what can be inferred even from
accurate context.
The omitted information includes room footprints for room-size estimation,
motion between stored poses for path-length estimation, attributes that
distinguish target instances for 3D grounding, and target objects or
sufficiently precise measurements for dimensional measurement.

Figure~\ref{fig:gt_residual_breakdown} shows that these modes recur across
eight subtasks, with 52.5--99.1\% of their GT-context failures matching
the diagnostic rules.
Section~\ref{sec:diagnostic_results} examines the task-level results and
Appendix~\ref{sec:qualitative_evidence} provides the detailed examples.

\finishwrap


\section{\methodname: Composable Reasoning Operators and Spatial Skills}
\label{sec:method}

\begin{figure}[t]
    \centering
    \includegraphics[width=\textwidth]{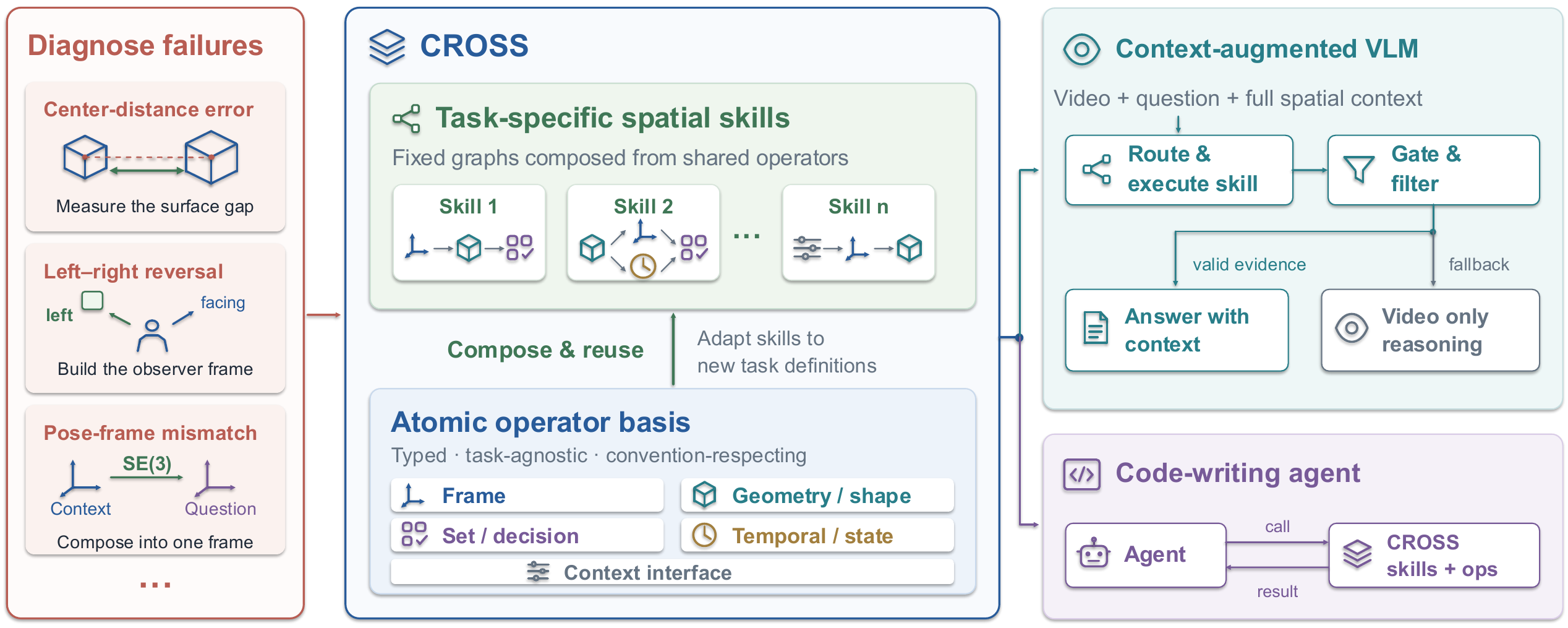}
    \caption{Overview of \methodname{}.
    Overview of CROSS. Operators derived from diagnosed failures (left) are composed into task-specific skills (middle) that serve two interfaces (right): context-augmented VLMs receive gated and filtered skill outputs or fall back to video-only reasoning, and code-writing agents call the skills and operators directly.
    }
    \label{fig:cross_overview}
\end{figure}

The controlled diagnosis reveals that models repeatedly reconstruct the same
spatial procedures and repeatedly make the same mistakes: they select the
wrong measurement, mix coordinate frames, reverse directional conventions, or
fail to update state.
\methodname{} avoids this inefficient and fragile rediscovery by assigning
these recurring computations to verified atomic operators and composing them
into task-specific spatial skills.
Figure~\ref{fig:cross_overview} gives an overview.

\subsection{Applying \methodname{} to augmented context}

\textbf{Applying spatial skills and operators.}
Let $c$ denote the structured spatial context and let $z_0=(q,c)$ be the
initial typed state for question $q$.
A lightweight router selects the corresponding spatial skill, whose $i$-th
atomic operator $o_i$ updates the state as $z_i=o_i(z_{i-1})$ for $i=1,\ldots,n$.
The final state $z_n$ contains the derived result, its supporting evidence,
and validity information.
This division lets the pipeline reuse a convention-specified procedure instead
of asking the answering model to reconstruct a spatial algorithm for every
question; Sections~\ref{sec:cross_operators} and~\ref{sec:cross_skills} detail
the operator basis and its skill compositions.

\textbf{Gating and filtering the augmented context.}
Because perception tools may fail to reconstruct the entities or geometry
required by a question, we apply a shared deterministic safeguard $g$ after
operator execution:
\begin{equation}
    \widetilde{c}=g(c,z_n)=
    \begin{cases}
        \operatorname{filter}(c,z_n), & \text{if the derivation is valid},\\
        \varnothing, & \text{otherwise}.
    \end{cases}
    \label{eq:gate_filter}
\end{equation}
For a valid derivation, the filter retains the derived result and its
question-relevant support while removing unrelated entities and geometry.
If the derivation is rejected, $\widetilde{c}=\varnothing$ and the model falls
back to video-only reasoning.
Thus $g$ prevents detectably unreliable results from entering the prompt and
compresses accepted context without changing the preceding computation.

\subsection{Diagnosis-derived atomic operator basis}
\label{sec:cross_operators}

\begin{wraptable}{r}{0.44\textwidth}
\vspace{-\intextsep}
    \centering
    \footnotesize
    \setlength{\tabcolsep}{3pt}
    \caption{Atomic operator families.
    }
    \label{tab:operators}
    \begin{tabular}{@{}p{0.34\linewidth}p{0.60\linewidth}@{}}
        \toprule
        Family & Role \\
        \midrule
        Frame & \raggedright Align bases and poses. \tabularnewline
        Geometry/\newline shape & \raggedright Measure surfaces, extents, and floor area. \tabularnewline
        Set/decision & \raggedright Deduplicate, count, and match with margins. \tabularnewline
        Temporal/\newline state & \raggedright Integrate paths; update position and heading. \tabularnewline
        \bottomrule
    \end{tabular}
\end{wraptable}

The systematic failure modes in Section~\ref{sec:diagnostic_main} determine
the operator basis in Table~\ref{tab:operators}.
Wrong-measurement errors motivate geometry and temporal operators;
coordinate-frame and handedness errors motivate frame operators; and repeated
left--right and route-state errors motivate local-sector and state-update
operators.
The resulting operators have single responsibilities and documented
conventions, rather than encoding one patch per benchmark question.
The full operator inventory and family responsibilities appear in
Appendix~\ref{sec:cross_operators_appendix}.

\finishwrap

\subsection{Composing operators into spatial skills}
\label{sec:cross_skills}

For task $k$, we define its spatial skill as the ordered operator sequence
$s_k=(o_{k,1},\ldots,o_{k,n_k})$.
Context-augmented inference appends the shared safeguard $g$ to this sequence:
\begin{equation}
    s_k^{\mathrm{ctx}}:\quad
    z_0 \xrightarrow{o_{k,1}} z_1 \xrightarrow{o_{k,2}} \cdots
    \xrightarrow{o_{k,n_k}} z_{n_k} \xrightarrow{g} \widetilde{c}.
    \label{eq:skill_composition}
\end{equation}
The operators encode reusable spatial primitives, whereas $s_k$ fixes their
task-specific roles and execution order; code-writing deployment reuses this
operator sequence without the context-specific safeguard.
Adapting \methodname{} to a new task therefore requires mapping its definition
to an existing skill or recombining the same operator basis.

For example, the surface-distance skill aggregates reliable entity geometry
before measuring the closest boundary gap, directly addressing center-distance
substitution.
The egocentric-direction skill constructs an observer-centered basis and
projects the target into it, making perspective and handedness explicit.
For rigid-pose alignment, the skill composes camera motion and the question's
initial pose in a common $SE(3)$ frame before matching answer candidates.
The complete skill graphs and fallback conditions appear in supplementary
Table~\ref{tab:skills}.


\begin{table}[t]
    \centering
    \footnotesize
    \setlength{\tabcolsep}{2.0pt}
    \definecolor{diagnosticgain}{RGB}{34,110,65}
    \newcommand{\crossgain}[2]{#1\,{\scriptsize\textcolor{diagnosticgain}{(+#2)}}}
    \caption{Diagnosis and CROSS gains on \revsitiny{} and \restitiny{}.
    Both context and code-writing arms use Qwen3-VL-32B-Thinking. Green parentheses show gains over the paired baseline.
    }
    \label{tab:diagnostic}
    \begin{tabular}{@{}lrr*{3}{r>{\columncolor{blue!6}}r}@{}}
        \toprule
        & & Video & \multicolumn{2}{c}{Predicted context} &
        \multicolumn{2}{c}{GT context} & \multicolumn{2}{c}{Code writing} \\
        \cmidrule(lr){3-3}\cmidrule(lr){4-5}\cmidrule(lr){6-7}\cmidrule(l){8-9}
        Subtask & $n$ & CoT & Base & $+$\methodname{} & Base & $+$\methodname{} &
        SpatialClaw & $+$\methodname{} \\
        \midrule
        \multicolumn{9}{@{}l}{\textit{\revsitiny{}}} \\
        \quad Object counting & 118 & 58.2 & 46.2 & \crossgain{58.2}{12.0} & 93.1 & \crossgain{94.8}{1.7} & 42.4 & \crossgain{45.4}{3.0} \\
        \quad Absolute distance & 239 & 54.5 & 48.1 & \crossgain{63.5}{15.4} & 58.4 & \crossgain{100.0}{41.6} & 44.8 & \crossgain{49.6}{4.8} \\
        \quad Object size & 215 & 67.9 & 54.6 & \crossgain{67.9}{13.3} & 100.0 & 100.0 & 37.1 & \crossgain{37.2}{0.1} \\
        \quad Room size & 30 & 55.0 & 62.0 & 55.0 & 56.7 & 56.0 & 31.3 & \crossgain{33.3}{2.0} \\
        \quad Relative distance & 111 & 53.2 & 56.8 & 52.3 & 91.0 & 86.5 & 49.6 & \crossgain{55.0}{5.4} \\
        \quad Relative direction & 53 & 35.9 & 39.6 & \crossgain{62.3}{22.7} & 52.8 & \crossgain{100.0}{47.2} & 28.3 & \crossgain{62.3}{34.0} \\
        \quad Route planning & 57 & 36.8 & 38.6 & \crossgain{40.4}{1.8} & 29.8 & \crossgain{49.1}{19.3} & 33.3 & \crossgain{38.6}{5.3} \\
        \cmidrule(l){1-9}
        \quad Average & 823 & 55.9 & 50.0 & \crossgain{60.4}{10.4} & 76.2 & \crossgain{92.3}{16.1} & 40.7 & \crossgain{45.9}{5.2} \\
        \midrule
        \multicolumn{9}{@{}l}{\textit{\restitiny{}}} \\
        \quad Speed \& acceleration & 115 & 20.9 & 59.1 & \crossgain{61.7}{2.6} & 97.4 & \crossgain{100.0}{2.6} & 56.5 & 55.7 \\
        \quad Pose estimation & 142 & 24.7 & 27.5 & \crossgain{70.4}{42.9} & 66.2 & \crossgain{94.4}{28.2} & 66.2 & \crossgain{74.7}{8.5} \\
        \quad Displacement \& path length & 187 & 22.5 & 35.8 & \crossgain{39.6}{3.8} & 71.1 & \crossgain{75.4}{4.3} & 31.0 & \crossgain{38.0}{7.0} \\
        \quad Egocentric orientation & 102 & 17.7 & 40.2 & \crossgain{45.1}{4.9} & 97.1 & \crossgain{100.0}{2.9} & 37.3 & 37.3 \\
        \quad 3D video grounding & 177 & 17.5 & 20.3 & 20.3 & 74.6 & \crossgain{75.1}{0.5} & 16.4 & \crossgain{20.3}{3.9} \\
        \quad Dimensional measurement & 52 & 28.9 & 28.9 & 25.0 & 55.8 & \crossgain{59.6}{3.8} & 40.4 & \crossgain{42.3}{1.9} \\
        \cmidrule(l){1-9}
        \quad Average & 775 & 21.3 & 34.3 & \crossgain{43.9}{9.6} & 77.3 & \crossgain{84.7}{7.4} & 39.4 & \crossgain{43.5}{4.1} \\
        \bottomrule
    \end{tabular}
\end{table}

\textbf{Code-writing deployment.}
We expose the operators as callable routines and each $s_k$ as a callable skill that an agent selects and invokes within its native execution loop, with units, frames, and task semantics fixed by the operator contracts.

\section{Experiments}
\label{sec:experiments}

\begin{figure}[t]
    \centering
    \includegraphics[width=0.9\textwidth,trim={8bp 9bp 8bp 60bp},clip]{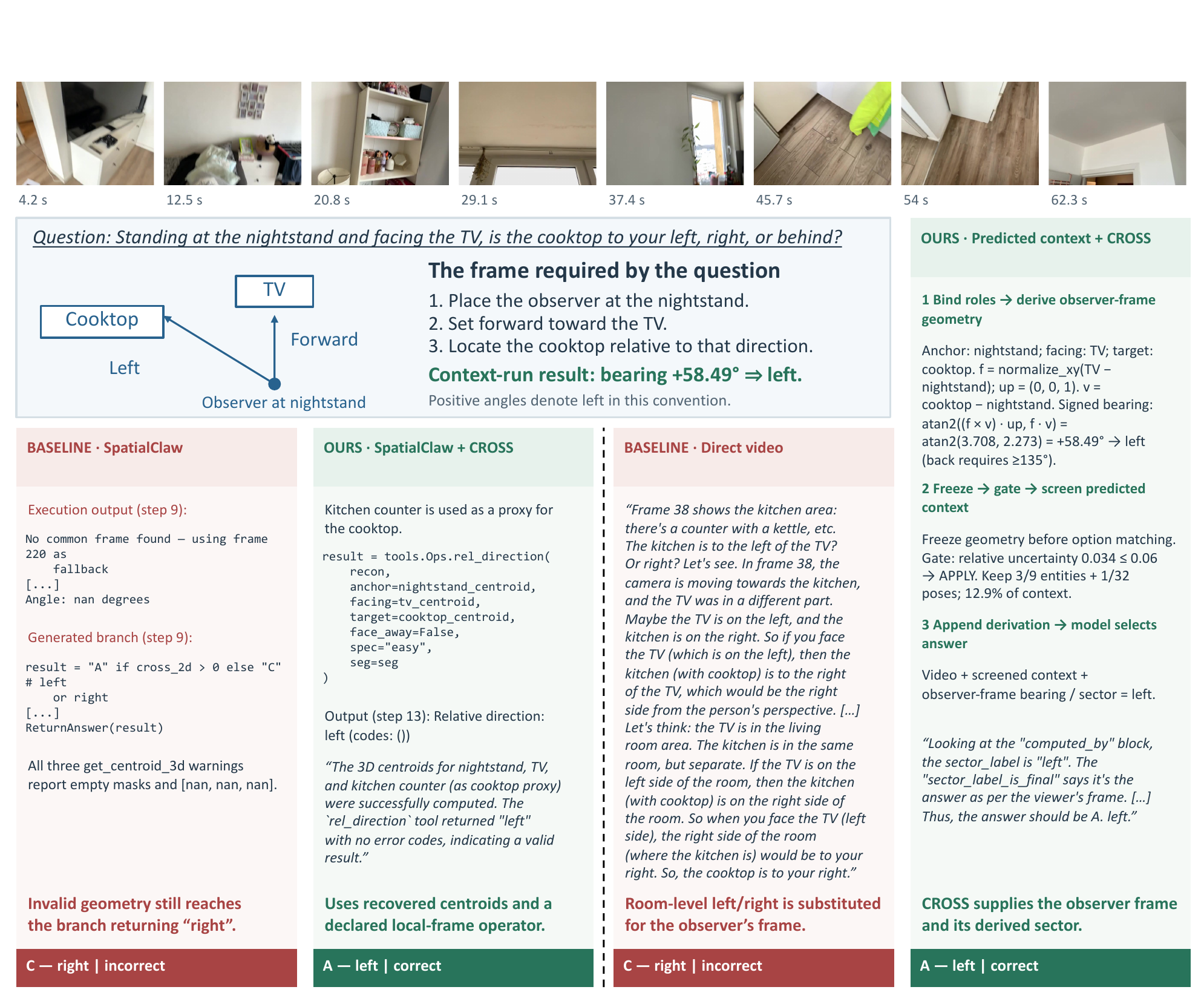}
    \caption{Relative-direction correction through code-writing (left pair) and
    context augmentation (right pair): both baselines answer right, whereas
    both \methodname{} runs answer left.}
    \label{fig:cross_baseline_corrections}
    \vspace{-1em}
\end{figure}

We first diagnose failures under controlled evidence and test their repair on
\revsitiny{} and \restitiny{}.
We then compare with prior work on ReVSI, VSI-Bench, and STI-Bench and test transfer to DSI-Bench.
Per-subtask effects, ablations, and limitations appear in Appendices~\ref{sec:ablations} and~\ref{sec:limitations}.

\subsection{Diagnostic experiments}
\label{sec:diagnostic_results}

\textbf{Scope.}
We conduct controlled diagnosis on two compact benchmark subsets using the
evidence and target validity checks in Appendix~\ref{sec:diagnostic_validity}.
Our \revsitiny{} evaluation uses 823 questions from ReVSI's released
tiny split \citep{zhang2026revsi}.
For dynamic spatial reasoning, we construct \restitiny{}, a 775-question
diagnostic subset of \resti{} \citep{sun2026restisourcegroundedauditrepair}
covering six tasks involving camera pose,
motion, time-indexed quantities, and metric measurement.
The diagnostic runs use dataset-specific frame budgets of 32 for
\revsitiny{} and 30 for \restitiny{}, held fixed across matched conditions.
Table~\ref{tab:diagnostic} reports the results; subset construction and
exclusion details are provided in Appendix~\ref{sec:diagnostic_scope_appendix}.

\textbf{Models and metrics.}
All context-based diagnostic experiments use Qwen3-VL-32B-Thinking with a shared decoding configuration; the code-writing arms evaluate SpatialClaw.
The four numerical \revsitiny{} tasks use the benchmark's mean relative
accuracy (MRA); its categorical tasks and all \restitiny{} tasks use
multiple-choice accuracy.
We report both in percent and differences in percentage points.

\textbf{Predicted context exposes different perception regimes.}
On \revsitiny{}, predicted context lowers the average from 55.9\% to 50.0\%,
chiefly on counting, absolute distance, and object size; on \restitiny{}, it
raises the average from 21.3\% to 34.3\% and improves five of six subtasks.
This contrast suggests that the current predictor recovers trajectory
structure more reliably than object identity and object-level metric geometry.

\textbf{Does GT context suffice for reasoning?}
GT context largely resolves tasks that require access to accurate entities or
camera geometry, but leaves errors in operation choice, frame conventions,
and evidence sufficiency.
Counting rises from 58.2\% to 93.1\% and object size from 67.9\% to 100.0\%;
relative distance reaches 91.0\%, whereas absolute distance remains at 58.4\%
because the model often substitutes center-to-center distance for the closest
surface gap.

Frame-sensitive tasks remain harder: relative direction reaches 52.8\%, with
left--right reversals and misinterpretations of ``facing away,'' which requires
reversing the anchor-to-reference vector by $180^{\circ}$.
Route planning falls from 36.8\% to 29.8\%, with left--right swaps accounting for
43 of 64 incorrect turn tokens.
This recurring signature does not by itself establish a failure to update heading.
Pose estimation reaches only 66.2\%, with errors matching omitted or incorrect
ego-to-world transformations.
By contrast, speed and egocentric orientation reach 97.4\% and 97.1\%.

The combined displacement-and-path-length row in Table~\ref{tab:diagnostic}
hides a further distinction: endpoint displacement reaches 99.2\% accuracy,
but path length reaches only 19.7\%, because sparse poses retain the endpoints
without preserving the motion between them.
Room size has an analogous representation limit: multiplying the dimensions
of a scene envelope, such as an axis-aligned bounding box (AABB), gives the
area of an enclosing rectangle rather than the queried floor polygon.
This rectangle can coincide with an axis-aligned rectangular room, but counts
the missing corner of an L-shaped room and can span multiple rooms.
Thus, accurate context can still be insufficient, even when the arithmetic is
correct.


\subsection{Effectiveness of \methodname{} on diagnosed failures}

\textbf{GT context.}
\methodname{} improves reasoning over the same GT context, as shown by the
paired scores and gain annotations in Table~\ref{tab:diagnostic}.
On \revsitiny{}, it raises the average from 76.2\% to 92.3\% (+16.1), with
the largest gains on absolute distance (+41.6) and relative direction (+47.2)---
the tasks most affected by the diagnosed wrong-measurement and frame errors.
The gains are not uniform: object size, relative distance, and room size do
not improve.
On \restitiny{}, the average rises from 77.3\% to 84.7\% (+7.4), led by pose estimation (+28.2).
These results show that accurate evidence alone does not remove the need for
convention-correct reasoning.


\begin{table}[t]
    \centering
    \scriptsize
    \setlength{\tabcolsep}{1.65pt}
    \caption{Detailed comparison on ReVSI.
    Numerical questions use MRA and multiple-choice questions use accuracy.
    Green parentheses show positive score-point gains over the corresponding
    SpatialClaw or direct-video baseline.}
    \label{tab:full-revsi}
    \resizebox{0.85\linewidth}{!}{%
    \begin{tabular}{@{}l*{9}{c}@{}}
        \toprule
        System & Frames & \multicolumn{4}{c}{Numerical questions (MRA)} &
        \multicolumn{3}{c}{Multiple-choice questions (Acc.)} & Avg. $\uparrow$ \\
        \cmidrule(lr){3-6}\cmidrule(lr){7-9}
        & & Obj. cnt. & Abs. dist. & Obj. size & Room size &
        Rel. dist. & Rel. dir. & Route plan & \\
        \midrule
        \multicolumn{10}{@{}l}{\textit{Baselines}} \\
        Chance (random) & All & -- & -- & -- & -- & 23.7 & 26.8 & 26.0 & -- \\
        Chance (frequency) & All & 52.2 & 40.1 & 17.4 & 20.9 & 25.8 & 31.9 & 30.2 & 31.4 \\
        \midrule
        \multicolumn{10}{@{}l}{\textit{Commercial models}} \\
        GPT-5.2 & 64 & 56.2 & 41.5 & 73.9 & 63.0 & 48.4 & 34.9 & 38.2 & 50.9 \\
        Gemini 3 Pro & 1 FPS & 60.1 & 54.7 & 79.3 & 51.9 & 68.1 & 56.0 & 56.4 & 60.9 \\
        \midrule
        \multicolumn{10}{@{}l}{\textit{Open-source general models}} \\
        InternVL3.5-38B & 64 & 43.8 & 60.6 & 70.2 & 58.4 & 57.4 & 45.9 & 42.7 & 54.1 \\
        Qwen3-VL-32B-Instruct & 64 & 46.9 & 65.0 & 70.4 & 55.8 & 53.8 & 34.0 & 47.3 & 53.3 \\
        \midrule
        \multicolumn{10}{@{}l}{\textit{Spatially specialized models}} \\
        SpaceR-7B (SG-RLVR) & 32 & 30.7 & 34.5 & 52.0 & 18.6 & 22.8 & 34.5 & 20.2 & 30.5 \\
        VST-7B-SFT & 4 FPS & 35.4 & 52.6 & 67.9 & 47.2 & 49.2 & 36.9 & 35.4 & 46.4 \\
        Cambrian-S-7B & 128 & 48.4 & 60.5 & 65.5 & 46.7 & 37.1 & 48.5 & 37.0 & 49.1 \\
        VLM3R-7B & 32 & 41.6 & 61.6 & 64.8 & 52.5 & 46.5 & 49.5 & 34.1 & 50.1 \\
        \midrule
        \multicolumn{10}{@{}l}{\textit{Published agentic methods}} \\
        S-Agent & 64 & 54.0 & 45.6 & 62.6 & 53.4 & 63.6 & 66.4 & 66.1 & 58.8 \\
        \midrule
        \multicolumn{10}{@{}l}{\textit{Our code-agent runs (Gemma-4-31B-FP8)}} \\
        SpatialClaw (DA3 $+$ SAM3) & 64 & 52.9 & 68.4 & 61.3 & 47.6 & 59.0 & 60.9 & 39.4 & 55.6 \\
        SpatialClaw $+$ \methodname{} & 64 & \benchmarkgain{53.4}{0.5} & \benchmarkgain{73.2}{4.8} & \benchmarkgain{65.1}{3.8} & \benchmarkgain{49.2}{1.6} & \benchmarkgain{62.2}{3.2} & \benchmarkgain{62.7}{1.8} & \benchmarkgain{46.5}{7.1} & \benchmarkgain{58.9}{3.3} \\
        \midrule
        \multicolumn{10}{@{}l}{\textit{Our context runs (Qwen3-VL-32B-Thinking)}} \\
        Direct (video only) & 64 & 63.5 & 58.6 & 68.7 & 50.6 & 56.6 & 41.6 & 51.6 & 55.9 \\
        \quad w/ pred. context & 64 & 45.5 & 50.1 & 55.8 & 69.2 & 59.2 & 42.9 & 42.3 & 52.2 \\
        \quad w/ pred. context, \methodname{} & 64 & \benchmarkgain{64.5}{1.0} & \benchmarkgain{63.6}{5.0} & 67.8 & \benchmarkgain{52.6}{2.0} & \benchmarkgain{59.3}{2.7} & \benchmarkgain{64.3}{22.7} & 49.1 & \benchmarkgain{\textbf{60.2}}{4.3} \\
        \bottomrule
    \end{tabular}
    }
\end{table}

\textbf{Predicted evidence across inference interfaces.}
\methodname{} also improves inference when the spatial context is predicted rather than supplied by GT annotations.
On \revsitiny{}, noisy predicted context lowers the average from 55.9\% for direct inference to 50.0\%, but adding \methodname{} raises it to 60.4\% (+10.4 over predicted context), exceeding both baselines.
On \restitiny{}, predicted context raises the average from 21.3\% to 34.3\%, and \methodname{} further improves it to 43.9\% (+9.6).
The benefit extends beyond GT context to imperfect predicted evidence.

For code-writing, we expand SpatialClaw's toolbox with
\methodname{} tools rather than supplying an augmented context.
This raises SpatialClaw's average from 40.7\% to 45.9\% (+5.2) on \revsitiny{} and
from 39.4\% to 43.5\% (+4.1) on \restitiny{}.
The improvements in both interfaces support the reuse of the same reasoning operators through either context augmentation or agent tool calls.

\textbf{Correcting the same question through both interfaces.}
Figure~\ref{fig:cross_baseline_corrections} asks whether the cooktop is left, right, or behind when facing the TV from the nightstand.
Left pair: empty masks produce invalid centroids and \texttt{Angle: nan degrees}, yet SpatialClaw returns C (right).
The \methodname{} run recovers centroids and calls \texttt{rel\_direction} to obtain left, with a kitchen-counter proxy for the cooktop.
Right pair: direct video reasoning confuses room-level and observer-relative directions; predicted context with \methodname{} supplies a $+58.49^{\circ}$ bearing and yields A (left).
The agent prompts and geometry differ, so this selected success does not isolate the operator's causal effect.

\subsection{Main evaluation on ReVSI, VSI-Bench, and STI-Bench}

\textbf{ReVSI.}
ReVSI is our primary large-scale test because it covers the same failure taxonomy as the diagnostic subset while using the 64-frame setting.
Table~\ref{tab:full-revsi} presents all seven subtask columns alongside prior work \citep{zhang2026revsi,dai2026sagent}.
\methodname{} improves the ReVSI average from 55.9\% for direct inference
and 52.2\% for predicted context to 60.2\% with Qwen3-VL-32B-Thinking fixed.
Our score is numerically 1.4 points above S-Agent's reported 58.8\% and
0.7 points below Gemini 3 Pro's 60.9\% \citep{dai2026sagent,zhang2026revsi}.
For code-based inference with Gemma-4-31B-FP8 \citep{team2026gemma}, adding
\methodname{} raises SpatialClaw's average from 55.6\% to 58.9\% (+3.3).
\methodname{} exceeds the spatially specialized models listed in
Table~\ref{tab:full-revsi}, without adding a spatial fine-tuning stage.
Cross-system protocol differences are discussed in Appendix~\ref{sec:limitations}.

\newsavebox{\benchmarkmeasure}
\newlength{\benchmarkheight}
\newlength{\benchmarknaturalheight}
\begin{table}[tb]
    \centering
    \scriptsize
    \setlength{\tabcolsep}{2pt}
    \newcommand{\measurebenchmark}[1]{%
        \sbox{\benchmarkmeasure}{%
            \begin{minipage}{0.325\textwidth}%
                #1%
            \end{minipage}}%
        \setlength{\benchmarknaturalheight}{%
            \dimexpr\ht\benchmarkmeasure+\dp\benchmarkmeasure\relax}%
    }
    \setlength{\benchmarkheight}{0pt}
    \setlength{\extrarowheight}{0pt}
    \newcommand{\recordbenchmarkheight}[1]{%
        \measurebenchmark{#1}%
        \ifdim\benchmarknaturalheight>\benchmarkheight
            \setlength{\benchmarkheight}{\benchmarknaturalheight}%
        \fi
    }
    \recordbenchmarkheight{
\begin{tabular}{@{}>{\raggedright\arraybackslash}p{0.60\linewidth}r@{}}
    \toprule
    Method & Avg. $\uparrow$ \\
    \midrule
    \multicolumn{2}{@{}l}{\textit{Baselines}} \\
    Chance (frequency) & 34.0 \\
    \midrule
    \multicolumn{2}{@{}l}{\textit{Commercial models}} \\
    GPT-5.2 & 49.2 \\
    Gemini 3 Pro & 60.5 \\
    \midrule
    \multicolumn{2}{@{}l}{\textit{Open-source general models}} \\
    InternVL3.5-38B & 60.8 \\
    Qwen3-VL-32B-Instruct & 61.8 \\
    \midrule
    \multicolumn{2}{@{}l}{\textit{Spatially specialized models}} \\
    SpaceR-7B (SG-RLVR) & 43.5 \\
    VLM3R-7B & 60.9 \\
    VST-7B-SFT & 65.2 \\
    Cambrian-S-7B & 67.5 \\
    \midrule
    \multicolumn{2}{@{}l}{\textit{Training-free (ours)}} \\
    Qwen3-VL-32B-Thinking & 59.8 \\
    w/ pred. context, \methodname{} & \benchmarkgain{\textbf{61.4}}{1.6} \\
    \bottomrule
\end{tabular}
}
    \recordbenchmarkheight{
\begin{tabular}{@{}>{\raggedright\arraybackslash}p{0.60\linewidth}r@{}}
    \toprule
    Method & Acc. $\uparrow$ \\
    \midrule
    \multicolumn{2}{@{}l}{\textit{Baselines}} \\
    Chance (random) & 20.0 \\
    \midrule
    \multicolumn{2}{@{}l}{\textit{Commercial models}} \\
    GPT-4o & 35.2 \\
    Gemini 2.5 Pro & 41.4 \\
    \midrule
    \multicolumn{2}{@{}l}{\textit{Open-source general models}} \\
    Qwen2.5-VL-7B-Instruct & 32.3 \\
    Qwen2.5-VL-72B-Instruct & 40.7 \\
    InternVL2.5-78B & 38.5 \\
    VideoChat-Flash & 36.3 \\
    \midrule
    \multicolumn{2}{@{}l}{\textit{Spatially specialized models}} \\
    SpaceR-7B (SG-RLVR) & 38.7 \\
    \midrule
    \multicolumn{2}{@{}l}{\textit{Training-free (ours)}} \\
    Qwen3-VL-32B-Thinking & 41.6 \\
    w/ pred. context, \methodname{} & \benchmarkgain{\textbf{45.3}}{3.7} \\
    \bottomrule
\end{tabular}
}
    \recordbenchmarkheight{

    \begin{tabular}{@{}>{\raggedright\arraybackslash}p{0.60\linewidth}r@{}}
        \toprule
        Method & Acc. $\uparrow$ \\
        \midrule
        \multicolumn{2}{@{}l}{\textit{Full benchmark}} \\
        Gemini 2.5 Pro & 46.9 \\
        SpatialTrackerV2 & 42.4 \\
        PolyV & 58.7 \\
        \midrule
        \multicolumn{2}{@{}l}{\textit{SpatialClaw (1,000 examples)}} \\
        No tools & 45.3 \\
        SpaceTools & 43.7 \\
        Toolshed & 43.0 \\
        Single-pass code & 57.9 \\
        Structured tool-call & 58.4 \\
        SpatialClaw & 62.9 \\
        \midrule
        \multicolumn{2}{@{}l}{\textit{Our runs (full benchmark)}} \\
        SpatialClaw reproduction & 62.8 \\
        SpatialClaw $+$ \methodname{} & \benchmarkgain{\textbf{66.3}}{3.5} \\
        \bottomrule
    \end{tabular}
}
    \newcommand{\alignedbenchmark}[4]{%
        \measurebenchmark{#1}%
        \begin{minipage}[t]{0.325\textwidth}
            \vspace{0pt}
            \centering
            \caption{#2 results.}
            \label{#3}
            \setlength{\extrarowheight}{%
                \dimexpr(\benchmarkheight-\benchmarknaturalheight)/#4\relax}
            #1
        \end{minipage}%
    }
    \alignedbenchmark{}{VSI-Bench}{tab:vsibench}{20}\hfill
    \alignedbenchmark{}{STI-Bench}{tab:stibench}{16}\hfill
    \alignedbenchmark{}{DSI-Bench}{tab:generalization}{15}
    \par\smallskip
\end{table}

\textbf{VSI-Bench and STI-Bench.}
Tables~\ref{tab:vsibench} and~\ref{tab:stibench} compare our training-free method with published
systems on the released benchmarks
\citep{yang2025thinking,zhang2026revsi,li2025stibench,ouyang2025spacer}.
We use 64 video frames for VSI-Bench and at most 30 uniformly sampled frames
for STI-Bench.
On VSI-Bench, \methodname{} raises the average over eight subtasks from
59.8\% to 61.4\% (+1.6).
On STI-Bench, the average rises from 41.6\% to 45.3\% (+3.7), exceeding the listed
Gemini 2.5 Pro (41.4\%) and SpaceR (38.7\%) scores.

\subsection{Cross-benchmark transfer on DSI-Bench}
\label{sec:cross_benchmark_transfer}

Table~\ref{tab:generalization} tests transfer through the code-writing
interface on the full DSI-Bench.
Published rows, including SpatialClaw's results on a fixed-seed subset of
1,000 questions, are external context only
\citep{zhang2025dsibench,wu2026polyv,cho2026spatialclaw}.
Our full-set SpatialClaw reproduction obtains 62.8\%, close to the 62.9\% reported
on that subset.
With the atomic operators frozen and composed into skills for DSI-Bench tasks,
SpatialClaw$+$\methodname{} reaches 66.3\%, a 3.5-point gain over our
reproduction.
This result supports transfer of the operator basis through the code-writing
interface; it does not establish transfer through context augmentation on
DSI-Bench.

\section{Conclusion}
\label{sec:conclusion}
In this paper, we show that accurate spatial evidence alone does not make video spatial reasoning reliable: with ground-truth context, models still select the wrong measurement or reference frame, while other failures stem from perception or missing information.
This diagnosis motivates \methodname{}, a training-free library of composable geometric operators and task-specific spatial skills that make these computations explicit and verifiable. 
Across five benchmarks and two inference interfaces, \methodname{} consistently improves spatial reasoning, including transfer to unseen tasks through a shared operator basis.
The results suggest a composable path toward spatial intelligence: improve perception and spatial representations to recover sufficient evidence, while using explicit, reusable operators to reason over that evidence correctly. 
More broadly, separating \emph{what spatial evidence is recovered} from \emph{how it is computed} offers a practical framework for building more reliable and transferable spatial reasoning systems, which we hope will inspire future research.



\subsection*{AI use statement}
We used generative AI assistants in four ways.
For writing, they drafted parts of sections, which the authors then revised,
and polished wording and LaTeX formatting throughout the manuscript.
For retrieval and discovery, they searched for and summarized related work,
and the authors checked the cited papers and bibliography entries against
their primary sources.
For research ideation and execution, they helped discuss experimental designs
and wrote and debugged code for experiments, analyses, and figures; the authors
verified this code and the resulting numbers against the logged model outputs.
The vision-language models evaluated in this paper are objects of study, not
writing aids.
The authors take full responsibility for the content of this paper, including
text, claims, and artifacts produced with AI assistance.

\bibliography{references}
\bibliographystyle{iclr2027_conference}

\clearpage
\appendix
\makeatletter\renewcommand\@fb@secFB{}\makeatother
\section*{Overview of the Appendix}

\begin{itemize}[leftmargin=*,topsep=2pt,itemsep=1pt]
    \item Appendix~\ref{sec:diagnostic_scope_appendix}: diagnostic subsets,
    implementation details, prompts, evidence and target checks, and the
    GT-context failure audit.
    \item Appendix~\ref{sec:trace_appendix}: coordinate contract, construction,
    and an example of the structured spatial context.
    \item Appendix~\ref{sec:cross_skills_appendix}: the operator inventory,
    skill graphs, routing and gating, and SpatialClaw integration of
    \methodname{}.
    \item Appendix~\ref{sec:ablations}: per-subtask effects of \methodname{} and
    ablations on VSI-Bench.
    \item Appendix~\ref{sec:limitations}: limitations.
    \item Appendix~\ref{sec:qualitative_evidence}: diagnostic and correction
    examples.
\end{itemize}

\section{Evaluation Setup and Failure Audit}
\label{sec:diagnostic_scope_appendix}

\subsection{Diagnostic subsets}

\textbf{\revsitiny{}.}
We draw from the 1,093-question tiny split released by ReVSI
\citep{zhang2026revsi} and evaluate the 823 questions for which all diagnostic
arms are available, after removing five questions from the corrupted scene
\texttt{d755b3d9d8}.
The questions span ScanNet, ScanNet++, and ARKitScenes and use 32 video frames.

\textbf{\restitiny{}.}
\restitiny{} is our diagnostic subset of \resti{}
\citep{sun2026restisourcegroundedauditrepair}, not an official split.
It keeps the six ScanNet indoor tasks whose source 3D annotations and camera
trajectories support GT-context construction, excludes scenes
\texttt{scene0002\_00} and \texttt{scene0069\_00} after the target checks
below, and contains 775 questions that use 30 video frames.

Within each subset, all arms share the questions, frame budget, and answer
parser, and averages weight questions equally.
The context and code-writing pipelines differ in reconstruction settings and
evaluation harness, so we compare base and \methodname{} runs only within each
interface.

\subsection{Implementation details}
\label{sec:implementation_details}

Table~\ref{tab:implementation_details} lists the inference settings.
Context-augmented runs use sampled decoding, and each arm is evaluated once.
SpatialClaw runs keep SpatialClaw's agent loop and add the \methodname{}
tools described in Appendix~\ref{sec:cross_spatialclaw_appendix}.

\begin{table}[!htb]
    \centering
    \small
    \setlength{\tabcolsep}{4pt}
    \caption{Inference settings for the two interfaces.}
    \label{tab:implementation_details}
    \begin{tabular}{@{}>{\raggedright\arraybackslash}p{0.16\linewidth}>{\raggedright\arraybackslash}p{0.38\linewidth}>{\raggedright\arraybackslash}p{0.40\linewidth}@{}}
        \toprule
        & Context augmentation & Code writing (SpatialClaw) \\
        \midrule
        Model & Qwen3-VL-32B-Thinking &
        Qwen3-VL-32B-Thinking (\revsitiny{}, \restitiny{});
        Gemma-4-31B-FP8 (ReVSI, DSI-Bench) \\
        Serving & vLLM, bf16 & vLLM \\
        Decoding & Sampling with temperature 1.0, top-$p$ 0.95, top-$k$ 20;
        up to 40,960 new tokens &
        Temperature 0.6; up to 32,768 (Qwen) or 131,072 (Gemma) tokens per call \\
        Video & Uniform frames: 32 (\revsitiny{}), 30 (\restitiny{}, STI-Bench),
        64 (ReVSI, VSI-Bench) &
        32 key frames shown to the model; up to 64 frames for reconstruction \\
        Perception & SAM3 and DA3 in streaming mode (predicted context;
        Appendix~\ref{sec:trace_appendix}) &
        SAM3 and DA3 in batch mode, called as agent tools \\
        Agent loop & -- & Up to 30 code steps with a 600\,s limit per step;
        planning enabled \\
        Scoring & Answers extracted with the VLMEvalKit spatial-benchmark
        matcher; MRA over thresholds 0.50--0.95 for numerical answers &
        SpatialClaw's evaluation harness with the same benchmark metrics \\
        \bottomrule
    \end{tabular}
\end{table}

\subsection{Prompts}
\label{sec:prompts}

Figure~\ref{fig:prompt_templates} gives the prompts of the three context arms.
The arms differ only in the blocks inserted before the question.
When the gate rejects a \methodname{} derivation, the question uses the Video
CoT prompt.

\begin{figure}[!htb]
    \centering
    \fbox{\begin{minipage}{0.96\linewidth}
    \footnotesize\raggedright
    \textbf{Video CoT}\\
    \texttt{You are answering a spatial reasoning question about an indoor
    scene shown in the video frames above.}\\
    \textit{[Time note]}\\
    \texttt{Question: \{question\}}\\
    \texttt{Options: \{options\}}\\
    \texttt{Think step by step, then provide your final answer in the format
    <answer>X</answer>.}

    \smallskip
    \textbf{Time note}\\
    \texttt{This question refers to the time interval from t = \{start\} s to
    t = \{end\} s, measured from the start of the video.}

    \smallskip
    \textbf{Context}\\
    After the Video CoT introduction, the prompt adds
    \texttt{Below is a structured representation of the scene, provided as
    additional context. Use it to inform your answer.}
    and the structured spatial context as JSON; the time note, question,
    options, and answer instruction follow.

    \smallskip
    \textbf{Context $+$ \methodname{}}\\
    After the filtered context, the prompt adds
    \texttt{The block below was computed deterministically from the structured
    scene representation above, using the coordinate contract and units it
    declares. It is not a guess and it is not derived from the answer options.}
    and the derived block as JSON.
    \end{minipage}}
    \caption{Prompt templates for the context arms.
    Typewriter text is verbatim and braces mark inserted fields.
    The time note appears only for time-scoped questions, with a single-instant
    variant, and ``an indoor scene'' becomes ``a scene'' for outdoor and
    tabletop sources.}
    \label{fig:prompt_templates}
\end{figure}

\subsection{Evidence and target checks}
\label{sec:diagnostic_validity}

Before attributing a GT-context error to reasoning, we recompute the requested
answer deterministically from the supplied context under the benchmark's
conventions.
If the answer is recoverable, the error counts toward the reasoning residual.
If the context omits a required variable, as with sparse trajectories or
missing room footprints, the example is context-insufficient.
If recomputation contradicts the released target, the example is invalid for
diagnosis.

\subsection{GT-context failure audit}
\label{sec:gt_context_audit}

Figure~\ref{fig:gt_residual_breakdown} audits the GT-context Base runs of
Table~\ref{tab:diagnostic} on the eight subtasks whose GT-context score is
below 90, with path length separated from endpoint displacement.
A failure is an incorrect multiple-choice answer or a numerical prediction with
MRA below 1.
For each failure, we evaluate the rules in Table~\ref{tab:gt_residual_audit}
on the supplied context, the prediction, and the reference answer, and assign
the first match.
A match shows consistency with a mechanism rather than a unique cause.

\begin{table}[!htb]
    \centering
    \small
    \setlength{\tabcolsep}{3pt}
    \renewcommand{\arraystretch}{1.12}
    \caption{Rules and denominators for Figure~\ref{fig:gt_residual_breakdown}.
    Counts are matching errors over all errors in the same GT-context Base run
    as Table~\ref{tab:diagnostic}; percentages are conditional on failure.
    Multiple rules within a row use first-match assignment.}
    \label{tab:gt_residual_audit}
    \begin{tabular}{@{}>{\raggedright\arraybackslash}p{0.23\linewidth}>{\raggedright\arraybackslash}p{0.54\linewidth}r@{}}
        \toprule
        Subtask & Matching rule & Matched/errors \\
        \midrule
        \multicolumn{3}{@{}l}{\textit{Wrong measurement}} \\
        Absolute distance & Prediction within 5\% of center-to-center distance & 225/227 \\
        \midrule
        \multicolumn{3}{@{}l}{\textit{Wrong frame or state}} \\
        Relative direction & Left--right mirror (14), reversed facing (5), or fixed scene axes (4) & 23/25 \\
        Route planning & Incorrect answer differs from the target only by left--right turn swaps & 21/40 \\
        Pose estimation & Nearest option to ego-as-world (22), unrotated displacement (6), or initial pose (4) & 32/48 \\
        \midrule
        \multicolumn{3}{@{}l}{\textit{Missing information}} \\
        Room size & Prediction within 5\% of scene length times width & 25/27 \\
        Path length & Selected option is nearest the chord sum of stored trajectory samples & 31/53 \\
        3D video grounding & Queried box absent (32), or present category has multiple instances (11) & 43/45 \\
        Dimensional measurement & Value not recoverable (10), or target category absent (6) & 16/23 \\
        \bottomrule
    \end{tabular}
\end{table}

\textbf{Wrong measurement and frame.}
The context's closest-surface gap reproduces all 239 absolute-distance targets,
yet 225 of the 227 errors equal the center distance within 5\%.
Deterministic geometry reproduces all 53 relative-direction labels.
In route planning, 21 of the 40 errors differ from the target only by
left--right turn swaps, which account for 43 of the 64 wrong turn tokens.
Correct frame composition recovers 138 of 142 pose answers; the three pose
hypotheses overlap (22, 15, and 13 matches), and the table reports first-match
counts.

\textbf{Missing information.}
In room size, 25 of the 27 errors equal the product of the scene length and
width within 5\%, and all 25 overshoot.
The path-length context keeps a median of 37 of 1,157 valid poses; the chord
sum over these poses recovers a median 77.8\% of the true length and selects
the correct option in only 12 of 66 questions.
Grounding accuracy is 12/44 when the queried box is absent from the context,
72/83 when it shares its category with other instances, and 48/50 when it is
unique.
Dimensional-measurement accuracy is 20/27 when the context can recover the
value and 9/25 otherwise.

\section{Structured Spatial Context}
\label{sec:trace_appendix}

\textbf{Coordinate contract.}
All geometric fields use one ego-anchored, gravity-aligned frame in meters.
The first valid camera center is the origin, $+Z$ points up, $+Y$ is the
horizontal projection of the first camera's viewing direction, and
$+X=Y\times Z$.
Camera poses are stored as $T_{\mathrm{ego}\leftarrow\mathrm{cam}}$, which
maps points from the OpenCV camera frame ($+X$ right, $+Y$ down, $+Z$ forward)
into the ego frame.

\textbf{Construction.}
GT context is compiled from source 3D annotations and camera trajectories.
Predicted context uses SAM3 masks and tracks, prompted with object categories
from the question, together with DA3 metric depth and camera poses; predicted
gravity is estimated from the camera up-vectors.
The predictor receives the video, the question, and the requested timestamps,
but no GT geometry.
Both sources share the schema, coordinate contract, and timestamp schedule,
and fields that reveal the source are removed before serialization.

\textbf{Example.}
Figure~\ref{fig:context_example} shows an abridged GT context for a
\revsitiny{} absolute-distance question, together with the derived block that
\methodname{} adds when the gate applies its derivation.
The context stores object centers and oriented extents, so the closest-surface
gap and the center distance are both recoverable; the derived block reports
the requested quantity.

\newsavebox{\contextexamplebox}
\begin{figure}[!htb]
    \centering
    \begin{lrbox}{\contextexamplebox}
    \begin{minipage}{0.96\linewidth}
    \scriptsize
\begin{verbatim}
{"scene_id": "41069021", "dataset": "arkitscenes",
 "coordinate_contract": {
   "frame_definition": "origin = first valid video-frame camera center; ...",
   "pose_matrix_rule": "T_A_from_B maps a homogeneous point expressed in B into A",
   "camera_pose": "T_ego_from_camera_opencv maps OpenCV-camera points into ego", ...},
 "scene_dimensions_lwh_m": [7.28, 3.99, 1.99],
 "entities": [
   {"id": "picture_27", "category": "picture", "aliases": ["picture", "wall picture"],
    "center_ego_m": [-0.92, 0.15, 0.73], "size_local_axes_m": [0.77, 0.05, 0.79],
    "axes_ego_from_object": [[0.94, -0.34, 0.0], [0.34, 0.94, 0.0], [0.0, 0.0, 1.0]]},
   {"id": "microwave_22", "category": "microwave", "aliases": ["microwave"],
    "center_ego_m": [2.96, 0.42, 0.37], "size_local_axes_m": [0.44, 0.28, 0.27], ...},
   ...],
 "camera_trajectory": {"full_valid_pose_count": 1878, "samples": [
   {"frame_index": 59, "time_s": 5.88, "position_ego_m": [0.07, -1.24, 0.86],
    "pose_ego_from_camera_opencv":
      [[0.86, 0.12, -0.49, 0.07], [0.50, -0.13, 0.85, -1.24],
       [0.04, -0.99, -0.17, 0.86], [0.0, 0.0, 0.0, 1.0]]},
   ...]}}
\end{verbatim}
    \vspace{-0.6em}
    \rule{\linewidth}{0.3pt}
\begin{verbatim}
{"computed_by": "deterministic geometry over the representation above",
 "skill": "abs_distance",
 "quantities": {"closest_surface_distance_m":
     {"value": 3.3268, "unit": "m", "frame": "ego", "uncertainty": 0.1165}},
 "roles": {"a": {"phrase": "wall picture", "matched_entities": ["picture_27"]},
           "b": {"phrase": "microwave", "matched_entities": ["microwave_22"]}},
 "center_distance_m_for_contrast": 3.9152, ...}
\end{verbatim}
    \end{minipage}
    \end{lrbox}
    \fbox{\usebox{\contextexamplebox}}
    \caption{Abridged GT context (top) and derived block (bottom) for
    ``Measuring from the closest point of each object, what is the direct
    distance between the wall picture and the microwave (in meters)?''
    The reference answer is 3.3\,m.
    Context values are rounded, and ``\ldots'' marks omitted fields and
    entries; the full context contains 19 entities and 32 trajectory samples.}
    \label{fig:context_example}
\end{figure}

\textbf{Predicted-context quality.}
Predicted camera positions deviate from GT by a median of 0.39\,m (90th
percentile 1.55\,m).
The median gravity tilt is $9.2^{\circ}$, but 59 of 404 scene records exceed
$30^{\circ}$, mostly because of rotational drift.
These errors bound what operators can recover from predicted context.

\section{\methodname{} Implementation}
\label{sec:cross_skills_appendix}

\subsection{Atomic operator inventory}
\label{sec:cross_operators_appendix}

Table~\ref{tab:operator_inventory} lists the operators in each family of
Table~\ref{tab:operators}.
The context safeguard $g$ (Equation~\ref{eq:gate_filter}) gates and filters
their outputs but is not itself an operator.

\begin{table}[t]
    \centering
    \small
    \caption{Full atomic operator inventory and family responsibilities.
    The shared safeguard $g$ controls whether derived evidence enters the
    augmented context and is separate from these spatial computations.}
    \label{tab:operator_inventory}
    \begin{tabular}{@{}p{0.18\textwidth}p{0.41\textwidth}p{0.34\textwidth}@{}}
        \toprule
        Family & Operators & Responsibility \\
        \midrule
        Frame & \raggedright \texttt{relative\_vector}, \texttt{build\_ego\_basis},
        \texttt{project\_local}, \texttt{relative\_se3}, \texttt{compose\_se3},
        \texttt{pose\_distance} & \raggedright Construct directed vectors and a handedness-aware
        forward/right/up basis, transport poses between declared frames, and score
        pose candidates. \tabularnewline
        Geometry/shape & \raggedright \texttt{clean\_geometry},
        \texttt{aggregate\_geometry}, \texttt{surface\_distance},
        \texttt{oriented\_extent}, \texttt{floor\_polygon},
        \texttt{polygon\_area} & \raggedright Robustly aggregate predicted evidence and compute
        the benchmark-defined surface, extent, or footprint quantity. \tabularnewline
        Set/decision & \raggedright \texttt{deduplicate}, \texttt{cardinality},
        \texttt{reduce\_set}, \texttt{rank\_with\_margin},
        \texttt{match\_candidate} & \raggedright Operate over instance sets and match a computed
        result to candidates only when its margin exceeds estimated error. \tabularnewline
        Temporal/state & \raggedright \texttt{endpoint\_displacement}, \texttt{path\_integral},
        \texttt{interval\_rate}, \texttt{classify\_sector},
        \texttt{advance\_agent} & \raggedright Bind motion to an interval, distinguish endpoint
        from path quantities, classify local turns, and persistently update route
        position and heading. \tabularnewline
        \bottomrule
    \end{tabular}
\end{table}

\FloatBarrier
\subsection{Task-specific skill graphs}

Table~\ref{tab:skills} gives the operator composition and main fallback
condition of each skill.
In context augmentation, each skill runs on the full structured context before
$g$; code-writing agents call the same operator sequence directly.

\begin{table*}[t]
\centering
\scriptsize
\caption{Subtask skills and principal validity failures.
Each operator sequence runs over the full structured spatial context and is
followed by gating and filtering in the context-augmented realization.}
\label{tab:skills}
\begin{tabular}{p{0.135\textwidth}p{0.525\textwidth}p{0.255\textwidth}}
\toprule
Skill & Fixed operator graph & Main fallback condition \\
\midrule
Counting & \texttt{deduplicate $\rightarrow$ cardinality} & Missing/merged objects or severe track fragmentation. \\
Absolute distance & \texttt{select/clean/aggregate $\rightarrow$ surface\_distance $\rightarrow$ reduce(min)} & A role is absent or geometric uncertainty is high. \\
Relative distance & Absolute-distance graph per candidate, then \texttt{reduce(min) $\rightarrow$ rank\_with\_margin} & Candidate coverage is incomplete or the rank margin is too small. \\
Relative direction & \texttt{relative\_vector $\rightarrow$ basis/project/classify} & Missing role, degenerate facing, or boundary ambiguity. \\
Object size & \texttt{select/clean/aggregate $\rightarrow$ oriented\_extent $\rightarrow$ reduce(max)} & Truncation or unstable multi-view extent. \\
Room area & \texttt{select\_region $\rightarrow$ clean $\rightarrow$ floor\_polygon $\rightarrow$ polygon\_area} & Only a global envelope or incomplete floor is available. \\
Route turns & \texttt{relative\_vector $\rightarrow$ basis/project/classify $\rightarrow$ advance\_agent} per waypoint, then \texttt{match\_candidate} & Ambiguous landmark or unspecified terminal heading. \\
Displacement & \texttt{endpoint\_displacement} & Interval endpoints or timestamps are missing. \\
Path length & \texttt{path\_integral} & The context is sparse or trajectory jitter exceeds the error budget. \\
Interval speed & \texttt{endpoint\_displacement $\rightarrow$ interval\_rate} & Interval binding fails or the quantity definition is inconsistent. \\
Pose estimation & \texttt{relative\_se3 $\rightarrow$ compose\_se3 $\rightarrow$ pose\_distance $\rightarrow$ rank} & Frames are incompatible or a queried pose is missing. \\
\bottomrule
\end{tabular}
\end{table*}

\subsection{Routing, gating, and filtering}
\label{sec:cross_gate_appendix}

\textbf{Routing.}
The router is rule-based.
It selects a skill from the benchmark's task label and uses fixed patterns to
extract the question's semantic roles, such as the anchor, facing, and target
objects, the route landmarks, or the queried time interval, together with the
answer convention.
Each role phrase is bound to context entities by exact category, synonym,
head-noun, or token-overlap matching; multiple matches are flagged as
ambiguous rather than resolved silently.

\textbf{Gating.}
Every operator returns its value with a unit, a frame, a declared error, and
validity codes, and the gate maps the resulting report to one of three
decisions.
It abstains when the context cannot represent the requested quantity, such as
a floor area without a room footprint or a path length from sparse poses.
It falls back when a role is unresolved, the geometry is degenerate or the
frames are incompatible, a direction lies within $5^{\circ}$ of a sector
boundary, the relative uncertainty exceeds a skill-specific limit, or the
margin between the best and second-best options is below a skill-specific
multiple of the declared error.
Otherwise, it applies the derivation.
Both abstention and fallback lead to video-only answering.
The skill-specific limits, and whether ambiguous roles or overlapping boxes
are rejected, are selected by grid search on a scene-disjoint development
split (30\% of scenes) of the predicted-context diagnostic runs, maximizing
expected accuracy when rejected questions are answered from video alone.
Table~\ref{tab:diagnostic} reports all diagnostic questions, including these
development scenes.

\textbf{Filtering and the derived block.}
For an applied derivation, the filter keeps the coordinate contract, the full
records of the entities bound to the question's roles, and the trajectory
samples at the start and end of the queried interval, or only the first sample
for static questions.
Counting keeps all entities, because restricting them to the queried category
would reveal the answer.
In predicted context, the retained entity records include reconstruction-support
fields, such as depth confidence and supporting frames, which the base context
omits.
The derived block reports the value, unit, frame, uncertainty, operator
sequence, and role bindings (Figure~\ref{fig:context_example}).
Answer options enter only the final matching step, after the estimate is fixed.

\subsection{Integration with SpatialClaw}
\label{sec:cross_spatialclaw_appendix}

For code writing, \methodname{} is added to SpatialClaw's tools as a Python
module next to its reconstruction and segmentation tools.
Each skill is a single function that takes the session's reconstruction,
segmentation masks, and role arguments, and returns the value with its operator
sequence, supporting evidence, and validity codes; the individual operators
remain callable for quantities that no skill returns.
The tool description asks the agent to verify masks and evidence before
measuring and to heed validity codes, and the agent decides whether to call a
skill.
For DSI-Bench, four additional skills---camera motion, object motion, distance
trend, and observer bearing change---compose the same frozen operators and
return the matching option text.

\section{Additional Results and Ablations}
\label{sec:ablations}

\subsection{Per-subtask effects of \methodname{}}
\label{sec:per_subtask_effects}

Tables~\ref{tab:diagnostic} and~\ref{tab:full-revsi} report per-subtask scores.
This section identifies the subtasks behind each average change, in percentage
points.

\textbf{GT context.}
After relative direction (+47.2) and absolute distance (+41.6), the largest
\revsitiny{} gain is route planning (+19.3, from 29.8\% to 49.1\%), consistent with
the left--right turn swaps among the GT-context failures
(Appendix~\ref{sec:gt_context_audit}).

\textbf{Predicted context.}
Pose estimation (+42.9, from 27.5\% to 70.4\%) and relative direction (+22.7,
from 39.6\% to 62.3\%) gain most.
Counting, object size, and room size match their video-only scores exactly
(58.2\%, 67.9\%, and 55.0\%), which indicates that the gate routed these questions to
video-only answering (Appendix~\ref{sec:cross_gate_appendix}).
Their changes relative to predicted context (+12.0, +13.3, and $-7.0$)
therefore reflect fallback rather than operator outputs.
Relative distance (56.8\% to 52.3\%) and dimensional measurement (28.9\% to 25.0\%)
also decrease.

\textbf{Code writing.}
Relative direction gains most (+34.0, from 28.3\% to 62.3\%).
Speed and acceleration decreases slightly (56.5\% to 55.7\%), and egocentric
orientation is unchanged (37.3\%).

\textbf{ReVSI.}
In the context runs, relative direction drives the average gain (+22.7, from
41.6\% to 64.3\%), whereas object size (68.7\% to 67.8\%) and route planning (51.6\% to
49.1\%) fall below direct inference.
With SpatialClaw, \methodname{} improves all seven subtasks, most on route
planning (+7.1) and absolute distance (+4.8).

\subsection{Ablations on VSI-Bench}
\label{sec:vsibench_ablations}

Table~\ref{tab:ablations} compares five configurations on VSI-Bench with
Qwen3-VL-32B-Thinking and 64 video frames.
The 32- and 64-frame contexts differ only in the temporal density of the
trajectory in the predicted context; the answering model always sees 64 video
frames.
Screened context removes question-irrelevant entities without applying
operators.

\noindent\begin{minipage}{\linewidth}
    \makeatletter\def\@captype{table}\makeatother
    \centering
    \scriptsize
    \setlength{\tabcolsep}{2pt}
    \caption{Ablations on VSI-Bench with 64 video frames.
    Average gives equal weight to eight subtasks.
    Numerical tasks use MRA and categorical tasks use accuracy (higher is better).
    Bold marks the best result in each column, including ties.}
    \label{tab:ablations}
    \begin{tabular*}{\linewidth}{@{\extracolsep{\fill}}l*{9}{r}@{}}
        \toprule
        Setting & \multicolumn{4}{c}{Numerical questions (MRA)} &
        \multicolumn{4}{c}{Multiple-choice questions (Acc.)} & Average $\uparrow$ \\
        \cmidrule(lr){2-5}\cmidrule(lr){6-9}
        & Obj. cnt. & Abs. dist. & Obj. size & Room size &
        Rel. dist. & Rel. dir. & Route plan & App. order & \\
        \midrule
        Direct & \textbf{60.8} & 50.2 & \textbf{73.9} & 61.3 & \textbf{57.2} & 57.5 & 48.7 & 68.7 & 59.8 \\
        32-frame context & 40.6 & 39.4 & 45.4 & \textbf{65.5} & 56.5 & 53.1 & 42.4 & 68.9 & 51.5 \\
        64-frame context & 41.9 & 39.0 & 46.0 & 64.7 & 56.8 & 53.4 & 38.7 & 69.2 & 51.2 \\
        Screened context & 42.3 & 41.2 & 46.9 & 57.0 & 55.9 & \textbf{62.6} & 46.6 & \textbf{69.7} & 52.8 \\
        \methodname{} & \textbf{60.8} & \textbf{53.4} & \textbf{73.9} & \textbf{65.5} & \textbf{57.2} & \textbf{62.6} & \textbf{49.2} & 68.7 & \textbf{61.4} \\
        \bottomrule
    \end{tabular*}
\end{minipage}\par\medskip

A denser trajectory does not help (51.2\% with 64 samples versus 51.5\% with 32),
and both context variants fall below direct inference (59.8\%).
Screening improves relative direction and route planning but hurts room size,
leaving the average at 52.8\%.
\methodname{} reaches 61.4\%, exceeding screened context by 8.6 points and
direct inference by 1.6, with gains on absolute distance, room size, relative
direction, and route planning.

\section{Limitations}
\label{sec:limitations}

The GT-context intervention shows whether accurate evidence is useful, but it
does not make perception and reasoning independent causal mechanisms.
Predicted context remains limited by open-vocabulary recall, monocular metric
scale, and rotational drift.
The gate is heuristic: its checks cannot catch every incorrect operator output,
and its room-scale assumptions may not transfer to outdoor scenes.
The operator basis does not cover acceleration or appearance order; the indoor
\resti{} subset contains no acceleration questions, and our diagnosis does not
study appearance order.
Spatial evaluation also depends on explicit target definitions: total 3D
rotation and gravity-relative yaw are both valid but measure different
capabilities.
Deterministic operators precompute part of the answer; their value lies in
enforcing conventions reliably, not in showing that the base model has acquired
an internal spatial algorithm.
Finally, published methods differ in backbone and frame budget, so their scores
provide descriptive context rather than matched comparisons, and our tables
report point estimates without significance tests.

\section{Qualitative Examples}
\label{sec:qualitative_evidence}

\begin{figure}[!tbp]
    \centering
    \includegraphics[width=\linewidth,height=0.78\textheight,keepaspectratio,
        trim={0 0 467 0},clip]{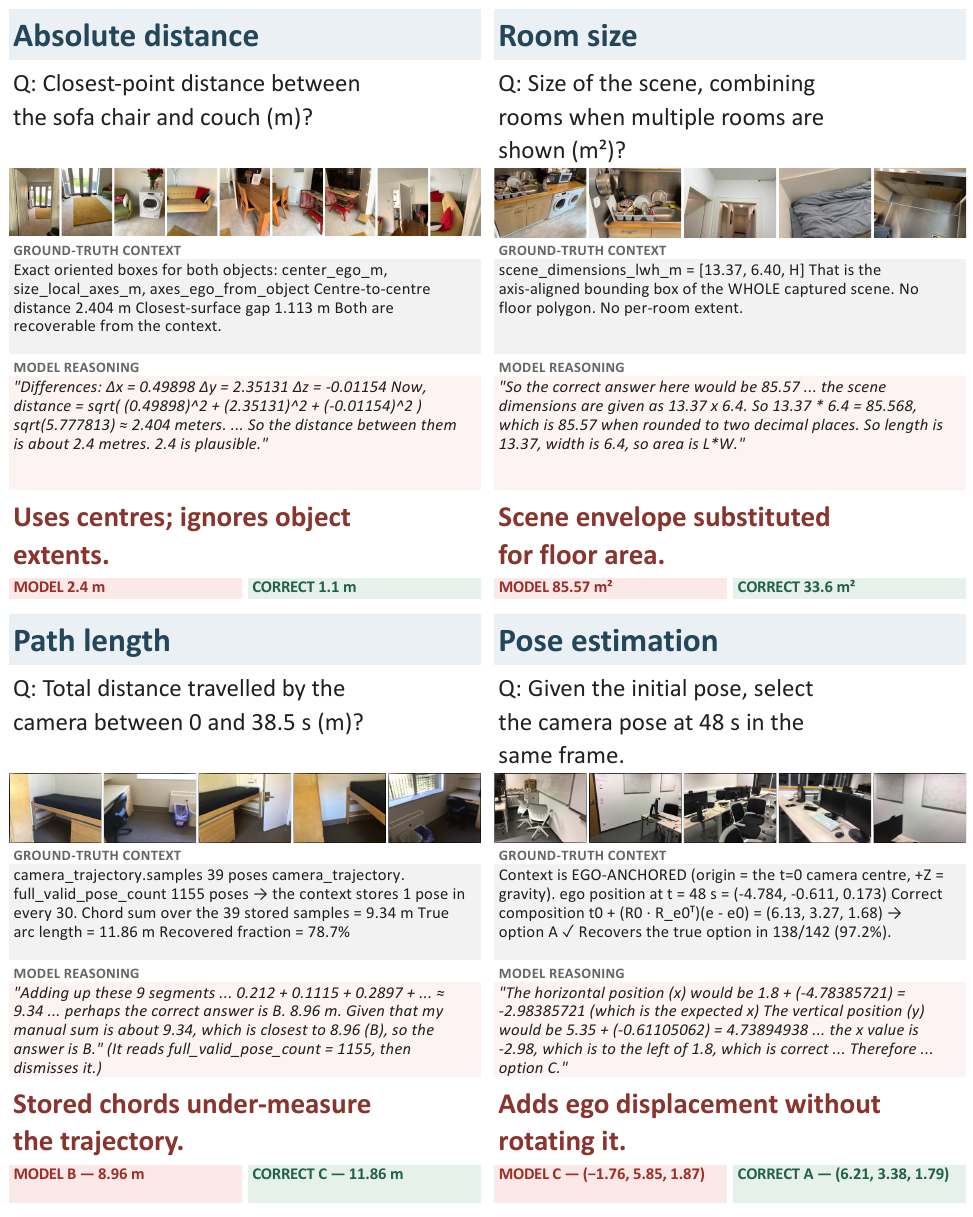}
    \caption{GT-context failures on absolute distance and room size (top) and
    path length and pose estimation (bottom): center distance used for the
    surface gap, the scene envelope used for floor area, a sparse trajectory
    that underestimates path length, and an ego displacement not rotated into
    the world frame.}
    \label{fig:gt_evidence_left}
\end{figure}

Figures~\ref{fig:gt_evidence_left} and~\ref{fig:gt_evidence_right} show one
GT-context failure for each subtask in Figure~\ref{fig:gt_residual_breakdown},
with \revsitiny{} cases in the upper rows and \restitiny{} cases in the lower
rows.
Each panel gives the question, video frames, supplied context, an excerpt of
the model's reasoning, the diagnosed failure, and the predicted and reference
answers.

\begin{figure}[p]
    \centering
    \includegraphics[width=\linewidth,height=0.88\textheight,keepaspectratio]{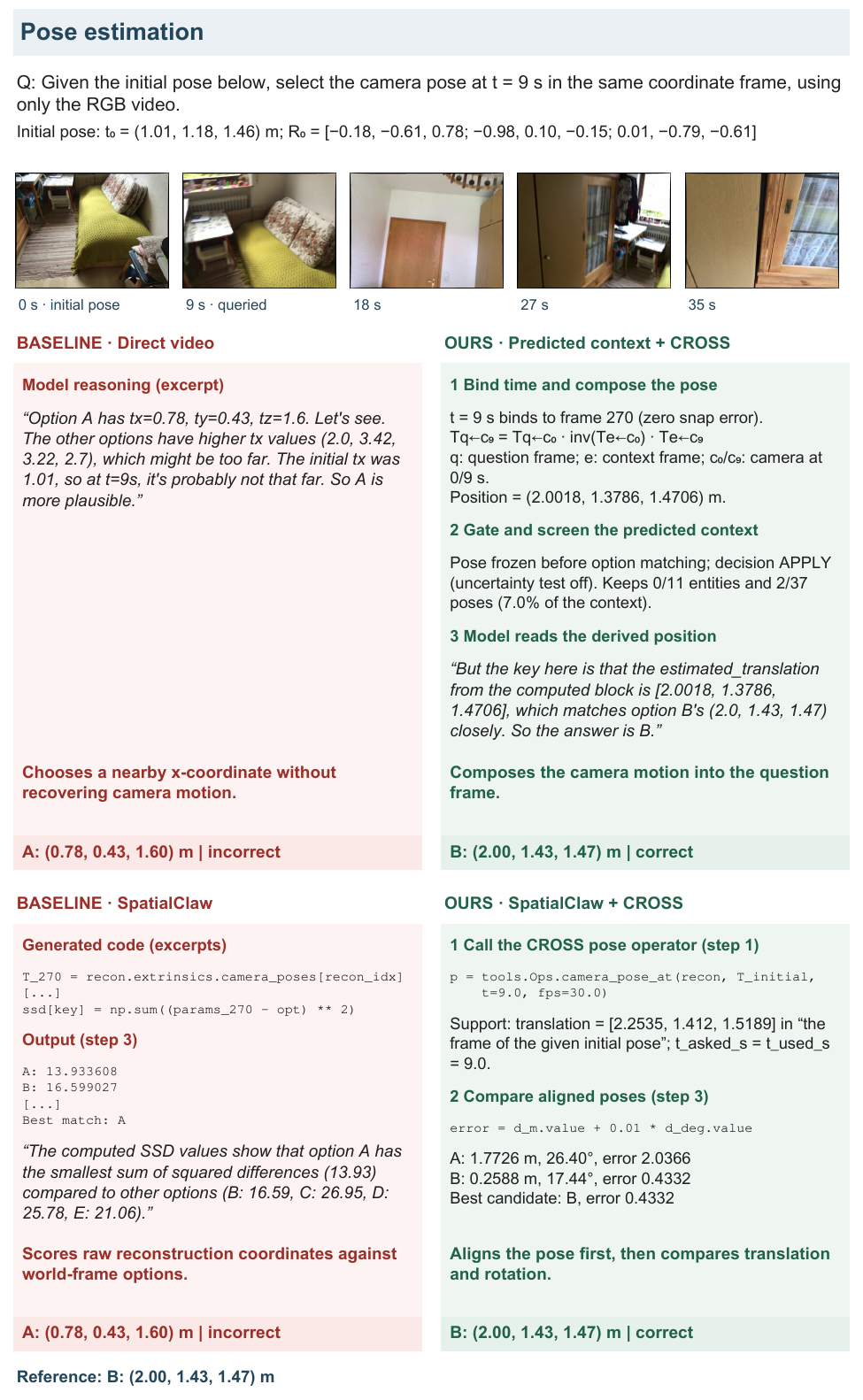}
    \caption{Pose-estimation correction through context augmentation (top) and
    SpatialClaw (bottom).
    Both baselines choose A; both \methodname{} runs compose the camera motion
    with the initial pose and choose B.}
    \label{fig:correction_pose_full}
\end{figure}

Figure~\ref{fig:correction_pose_full} follows one \restitiny{} pose question
across four arms.
Both baselines choose a wrong option; the SpatialClaw baseline compares
reconstructed poses with the question's world-frame options without aligning
them to the given initial pose.
Both \methodname{} runs compose the recovered camera motion with the initial
pose in a common frame and select the correct option.
The context and agent arms use separate reconstructions, and this context run
disables the gate's uncertainty test.

\begin{figure}[p]
    \centering
    \includegraphics[width=\linewidth,height=0.80\textheight,keepaspectratio]{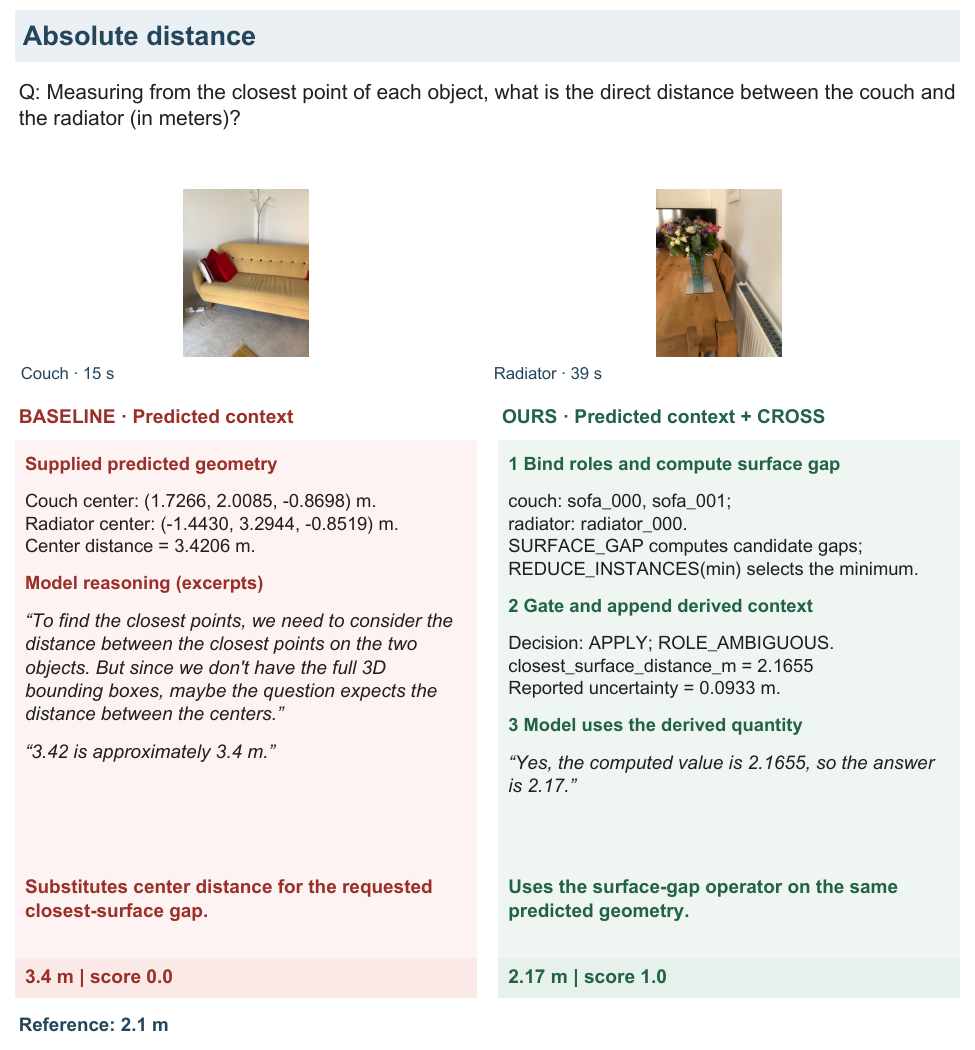}
    \caption{Absolute-distance correction.
    The baseline uses center distance and returns 3.4\,m; \methodname{}
    derives the closest-surface gap and returns 2.17\,m against a 2.1\,m reference.
    The gate retains the ambiguous couch-role warning while reducing candidate
    pairs by minimum gap.}
    \label{fig:correction_absolute_distance}
\end{figure}

\begin{figure}[p]
    \centering
    \includegraphics[width=\linewidth,height=0.80\textheight,keepaspectratio]{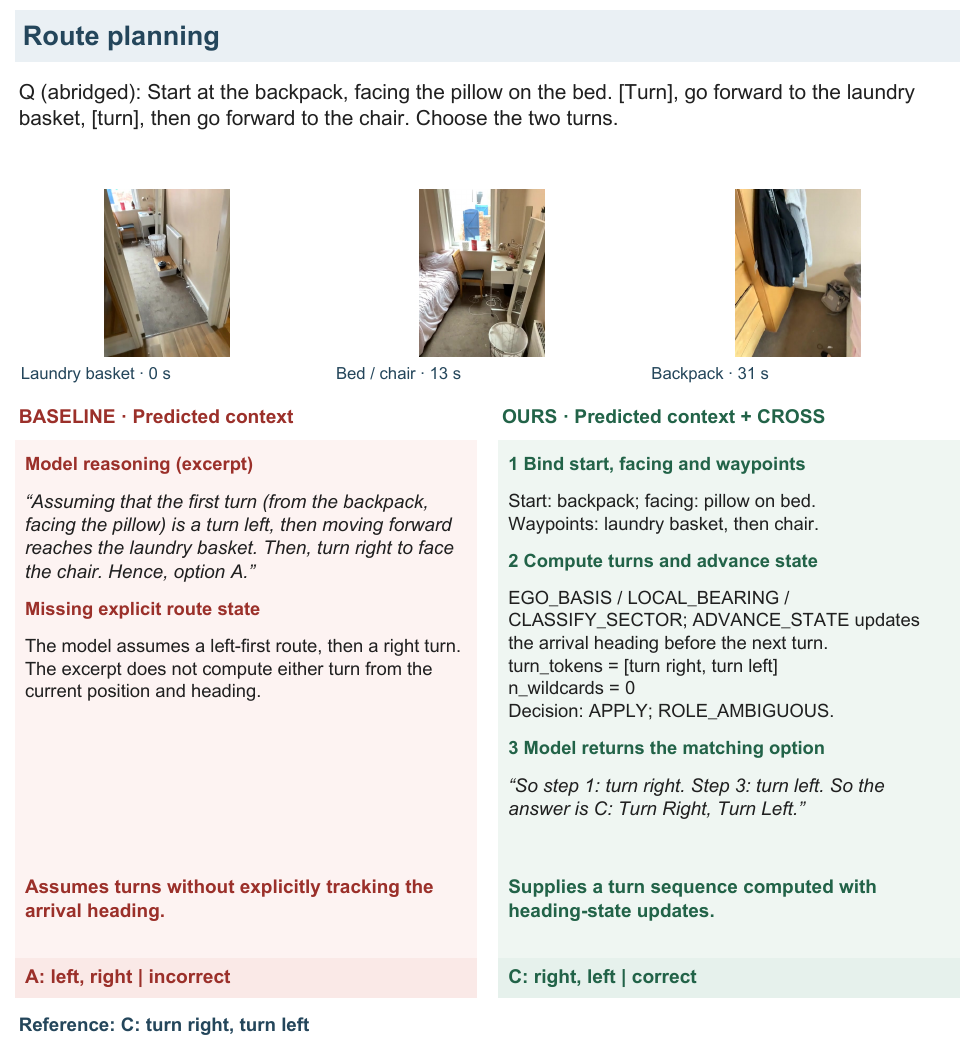}
    \caption{Route-planning correction.
    The baseline selects A (left, right); \methodname{} supplies the sequence
    computed with heading-state updates and the model selects C (right, left).
    Multiple bed and chair tracks remain in the bound roles.}
    \label{fig:correction_route_planning}
\end{figure}

\begin{figure}[p]
    \centering
    \includegraphics[width=\linewidth,height=0.80\textheight,keepaspectratio]{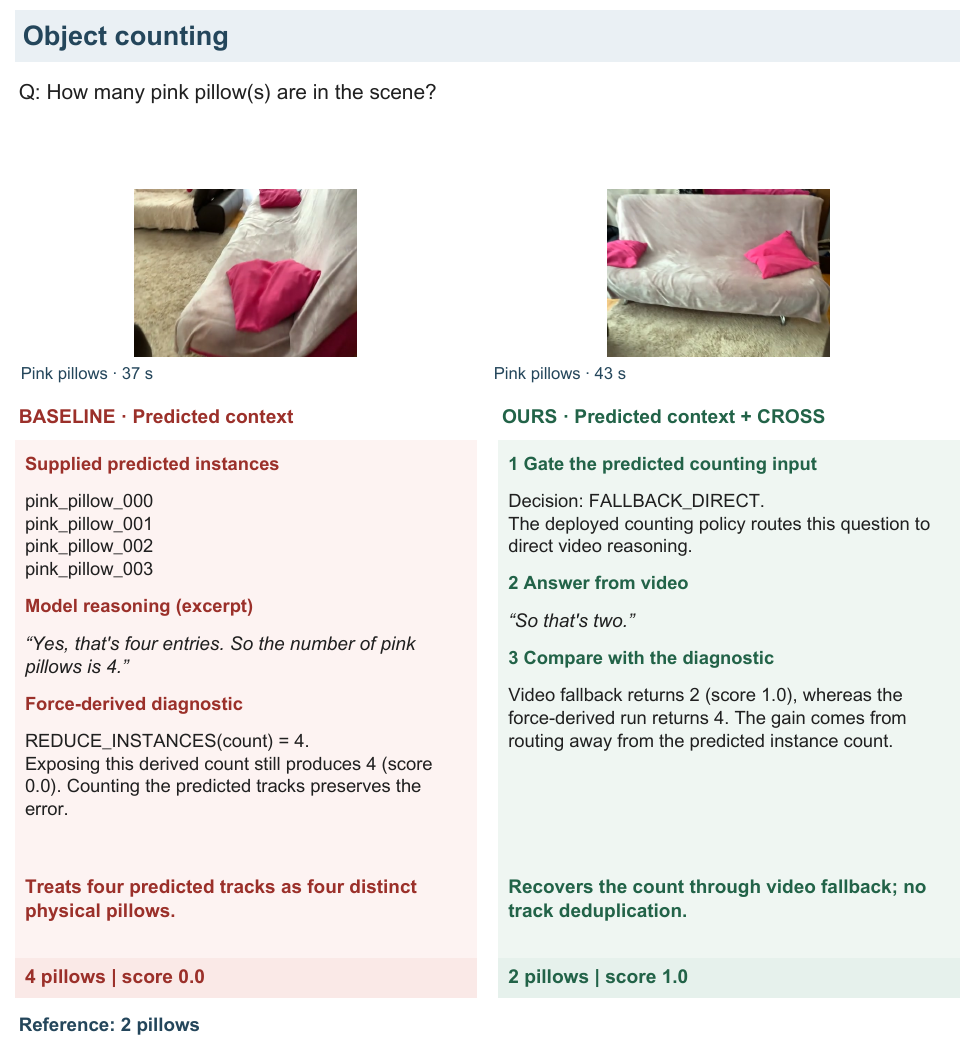}
    \caption{Object-counting recovery through video fallback.
    The baseline counts four predicted pink-pillow tracks; the deployed gate
    routes to video and the model returns the reference count of two.
    Forcing the derived count still returns four, distinguishing the routing
    benefit from instance reduction.}
    \label{fig:correction_object_counting}
\end{figure}

Figures~\ref{fig:correction_absolute_distance},
\ref{fig:correction_route_planning}, and~\ref{fig:correction_object_counting}
extend the paired layout to three other \revsitiny{} subtasks, comparing the
predicted-context baseline with \methodname{} on the same predicted context.
The distance case replaces center distance with a surface gap; the route case
supplies turns computed with heading-state updates.
The counting case instead recovers through video fallback: forcing the derived
count preserves the baseline's error.
These selected examples illustrate observed recoveries, not correction rates or
isolated operator effects.

\begin{figure}[p]
    \centering
    \includegraphics[width=\linewidth,height=0.78\textheight,keepaspectratio,
        trim={467 0 0 0},clip]{images/all_gt_residual_full_context_wide.pdf}
    \caption{GT-context failures on relative direction and route planning (top)
    and 3D grounding and dimensional measurement (bottom): a handedness
    reversal, a heading that is never updated, a referring expression that
    matches many entities, and answer options closer than the context's
    precision.}
    \label{fig:gt_evidence_right}
\end{figure}

\FloatBarrier

\end{document}